\documentclass[11pt]{article}
\usepackage[preprint]{acl}

\usepackage{fontspec}
\usepackage{xeCJK}
\usepackage{microtype}
\usepackage{latexsym}
\usepackage{graphicx}
\usepackage{amsmath}
\usepackage{amssymb}
\usepackage{booktabs}
\usepackage{tabularx}
\usepackage{enumitem}
\usepackage{placeins}

\newcommand{\method}{\textsc{MemDoctor}}

\newcommand{\stage}[1]{\textsc{#1}}

\title{
MemDoctor: Stage-wise Error Attribution and Repair for MLLM Agent Memory}

\ExplSyntaxOn
\tl_new:N \g_memdoctor_author_list_tl
\int_new:N \g_memdoctor_author_count_int
\NewDocumentCommand \PaperAuthor { m m }
  {
    \int_compare:nNnT { \g_memdoctor_author_count_int } > { 0 }
      {
        \int_compare:nNnTF { \int_mod:nn { \g_memdoctor_author_count_int } { 3 } } = { 0 }
          {
            \tl_gput_right:Nn \g_memdoctor_author_list_tl { \AND }
            \int_compare:nNnT { \g_memdoctor_author_count_int } > { 3 }
              { \addtolength{\titlebox}{4\baselineskip} }
          }
          { \tl_gput_right:Nn \g_memdoctor_author_list_tl { \And } }
      }
    \tl_gput_right:Nn \g_memdoctor_author_list_tl { #1 \\ #2 }
    \int_gincr:N \g_memdoctor_author_count_int
  }
\NewDocumentCommand \SetPaperAuthors { }
  { \exp_args:NV \author \g_memdoctor_author_list_tl }
\ExplSyntaxOff

\PaperAuthor{Ruiyang Jiang}{Shanghai Jiao Tong University}
\PaperAuthor{Xinyang Liao}{Xi'an Jiaotong University}
\PaperAuthor{Weikai Huang}{Shanghai Jiao Tong University}
\PaperAuthor{Yuqi Niu}{Shanghai Jiao Tong University}
\PaperAuthor{Junjie Zhou}{Xidian University}
\PaperAuthor{Yubin Zheng}{Shanghai Jiao Tong University}
\SetPaperAuthors

\begin{document}
\maketitle
\begin{abstract}
Multimodal large language models (MLLMs) now assist users across sessions of images and text, carried by an external memory of what was said and shown.
When such an assistant answers wrongly, the failure usually began earlier, in how a record was written, updated or retrieved, and the answer alone does not say where.
We formalize multimodal memory as four stages, \stage{Write}, \stage{Update}, \stage{Retrieve} and \stage{Answer}, and define nine error types across them.
A stage-resolved analysis on Mem-Gallery traces most failures to a source stage, and most of those lie before the answer.
Injecting known errors against fact-preserving control edits, we then follow each tainted record downstream.
Early errors survive to the answer as apparent evidence, and hurt most when no other record holds the fact.
A diagnosis that trusts the quoted evidence therefore blames the stage where the error surfaces, not the one where it began.
We propose \method{}, in which the MLLM that built the memory diagnoses its own failures.
It traces each needed fact backward through the four stages, accepts a diagnosis only when the cited evidence is found in memory, and repairs that stage before answering again.
On a 4B backbone, \method{} improves accuracy on both Mem-Gallery and MemLens, outperforming all evaluated 8B baselines, and the same loop also improves MuRAG and A-Mem.
\end{abstract}

\begin{figure}[!t]
  \centering
  \includegraphics[width=0.87\columnwidth]{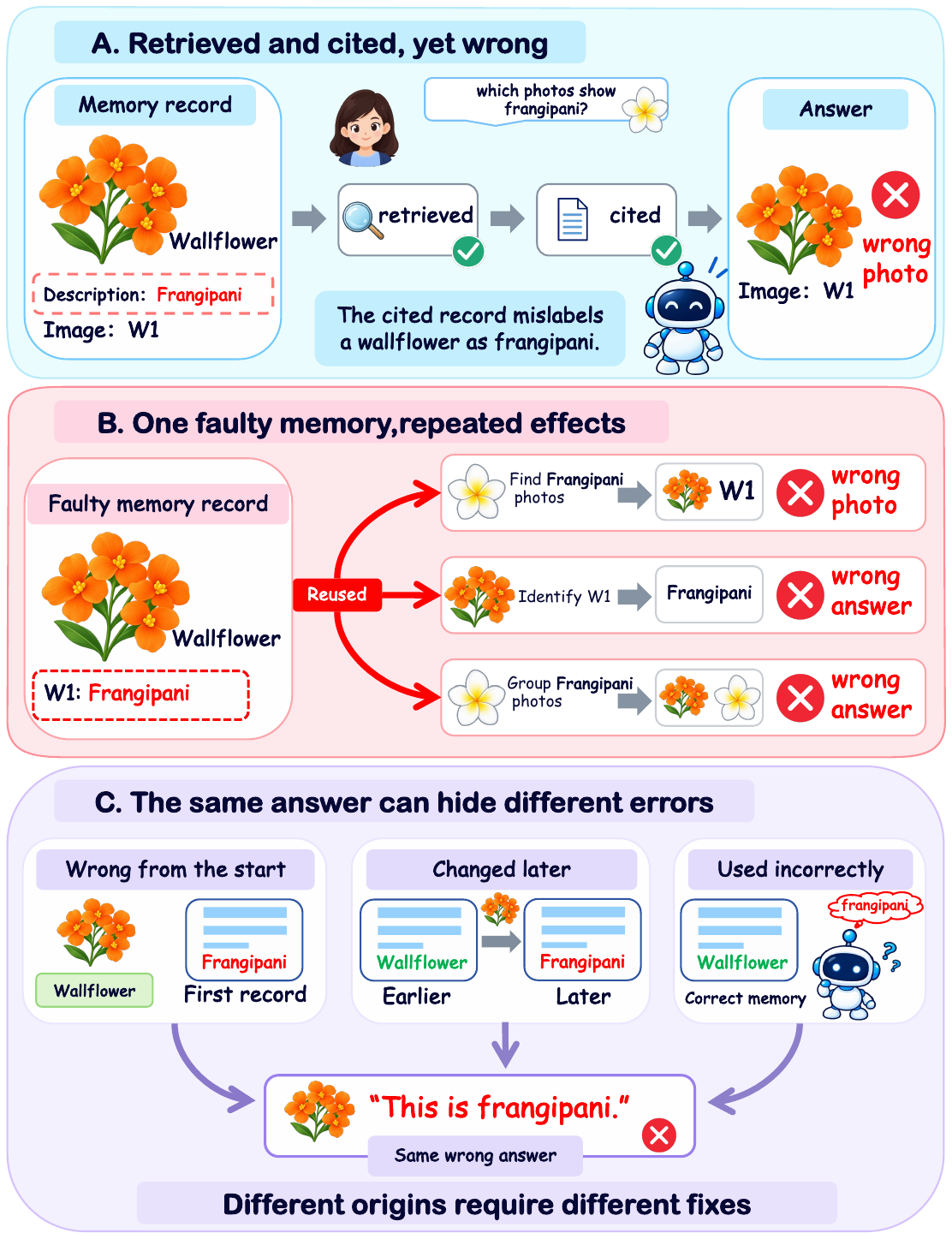}
  \caption{Memory errors pass as valid evidence, so an answer built on a faulty record looks grounded yet is wrong (A). They propagate to the later questions that read the same record (B). Since the same wrong answer can stem from writing, updating or answering, an error must be attributed to the stage where it began before it can be repaired (C). }
  \label{fig:memory_failure_overview}
\end{figure}

\section{Introduction}
\label{sec:intro}

Large language models (LLMs) can operate as agents with an agent harness that coordinates their interactions with environments and responses to feedback~\citep{achiam2023gpt4,yao2023react}.
Multimodal large language models (MLLMs) extend this interaction to visual inputs, allowing language instructions to be grounded in images~\citep{liu2023llava,bai2025qwen25vl,bai2025qwen3vl}.
As these interactions span multiple sessions, an MLLM agent must preserve relevant information~\citep{liu2024lost,levy2024same} from one session to answer questions in another~\citep{maharana2024locomo,bei2026memgallery,ren2026memlens,liu2026worldmemarena}, including not only what was said, but also what was shown, which entity a visual attribute belongs to, and how that entity changes over time.

From the perspective of how memory is constructed, maintained, and used~\citep{jiang2026anatomy}, memory-augmented answering can be broadly divided into four stages: \stage{Write} extracts evidence into records, \stage{Update} maintains them across interactions, \stage{Retrieve} selects evidence for a question, and \stage{Answer} uses that evidence to respond.
This decomposition raises a basic question: when an MLLM answers a memory-dependent question incorrectly, is the failure due to its memory, or to the model's inability to answer correctly even when sufficient evidence is available?
If the failure originates in memory, at which stage does it arise?
Answering these questions requires examining intermediate evidence and memory states, rather than the final response alone.

A faulty memory record can be retrieved and cited as valid evidence, carrying its error into later questions and updates (Figure~\ref{fig:memory_failure_overview}).
The same wrong answer can stem from writing, updating, or answering, each calling for a different repair.
For example, a record in Figure~\ref{fig:memory_failure_overview}A labels a wallflower image as frangipani; when asked which photos show frangipani, the agent retrieves and cites this record and incorrectly returns the wallflower image.
The memory appears faulty, but identifying which stage introduced the error requires tracing the record back through the memory pipeline.

\begin{figure}[t]
  \centering
  \includegraphics[width=0.85\linewidth]{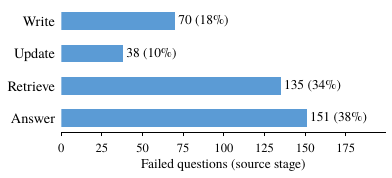}
  \caption{Source stage of the failures of the 4B system on Mem-Gallery.}
  \label{fig:stage_counts}
\end{figure}

We take Mem-Gallery, a benchmark of multimodal multi-session conversations, as the starting point for a stage-resolved analysis.
The initial analysis asks how many questions exhibit errors at each stage and whether later failures can be traced to earlier operations.
Stage-wise counts characterize where failures are observed; establishing their downstream effects additionally requires interventions.
On the 4B system of Table~\ref{tab:memgallery_main}, we trace 394 of its 467 failures to a source stage (Figure~\ref{fig:stage_counts}) by replacing each stage's output with a correct one and checking whether the answer is restored; the other 73 cannot be attributed.
Of the attributed failures, 62\% start before the answer stage, in retrieval (34\%), writing (18\%) or updating (10\%), yet every one surfaces in the answer.
Reading the failed trajectories stage by stage, we group the recurring errors into nine types: at \stage{Write}, \emph{omission}, \emph{distortion}, \emph{visual loss} (a fact visible only in the image) and \emph{visual misdescription} (the image is described as something similar but different); at \stage{Update}, \emph{update loss}; at \stage{Retrieve}, \emph{retrieval miss} and \emph{retrieval distraction} (the evidence is retrieved, but other retrieved records derail the answer); and at \stage{Answer}, \emph{misreasoning} and \emph{false recall} (a question the conversation never addresses is answered with invented content). 

We propose \method{}, a framework that diagnoses multimodal memory at two levels.
At the analysis level, it formalizes where an error starts and injects known errors to measure how they propagate through later operations.
At the system level, the MLLM that built the memory diagnoses and repairs its own failures.
After answering a question, and without access to the gold answer, the model works backward through these stages: it checks whether the retrieved records suffice; if not, whether the evidence exists elsewhere in memory but was not retrieved; if not, whether an earlier version of a record held it before an update; and otherwise whether it was never written.
Each check must be grounded in the memory state rather than in the model's own judgment: a diagnosis is accepted only when the evidence it cites can be located---in a record, in an earlier version of that record, or in the stored image---so the model cannot talk itself into a conclusion that the memory does not support.
A verified claim names a stage and a record; the model repairs that record accordingly, re-answers, and carries the outcome into the next round of reflection.

We make four contributions:
\begin{itemize}[leftmargin=*,itemsep=2pt,topsep=2pt]
  \item We formalize memory-augmented answering as four intervenable stages---\stage{Write}, \stage{Update}, \stage{Retrieve} and \stage{Answer}---and show that the failures of a multimodal memory system can be attributed to the stage where they begin rather than the stage where they surface.
  \item We examine how errors persist through later memory operations, separating their downstream effects from changes caused by replay itself.
  \item We propose \method{}, a self-attribution and repair framework in which the MLLM that built the memory diagnoses its own failures. 
  \item On a 4B backbone, \method{} outperforms every 8B baseline on both benchmarks. Attached as a plug-in to two external memory backends, the same diagnosis and repair loop improves both over their own baselines.
\end{itemize}

\section{Related Work}
\label{related_work}

Large language models (LLMs) support instruction following, reasoning, and response generation~\citep{achiam2023gpt4}.
Agent frameworks build on these capabilities to coordinate ongoing interactions with users and environments~\citep{yao2023react}.
As these interactions span multiple sessions, agents need external memory to preserve relevant information beyond their active context and retrieve it for later use.
MemGPT~\citep{packer2023memgpt} and MemOS~\citep{li2025memos} manage memory tiers, while A-Mem~\citep{xu2025amem} and Mem0~\citep{chhikara2025mem0} organize and revise persistent records.
LightMem~\citep{fang2026lightmem} and SimpleMem~\citep{liu2026simplemem} develop efficient consolidation mechanisms, and MAGMA~\citep{jiang2026magma} uses multiple graph views to organize retrieval.
For multimodal agents, memory must preserve not only what was said but also what was shown.
Multimodal memory systems address this requirement through episodic, semantic, and visual memories~\citep{long2026m3agent,yeo2026worldmm,liu2025memverse,feng2026m2a,bo2026vilomem}; Omni-SimpleMem~\citep{liu2026omnisimplemem} expands retrieved summaries into detailed or raw evidence, and V-Mem~\citep{kang2026vmem} routes retrieval by modality.

Prior work has explored how agents can inspect their own behavior and learn from failures, through reflection on task feedback~\citep{shinn2023reflexion,madaan2023selfrefine} and failure attribution within execution trajectories~\citep{zhang2025whowhen,zhang2026agentracer}.
These efforts motivate diagnosis that connects an unsuccessful outcome to the operations that produced it.
Within memory systems, MemTrace~\citep{deng2026memtrace} provides a closely related approach.
It records operation-level execution graphs and lets a stronger external model, given the gold answer, localize the decisive error through subgraph inspection or operation-log search; the attributions then guide prompt optimization.
We differ in four respects: 
(i) memory is multimodal: what an image showed must survive being written as text and carried through updates, which text-only analyses do not cover; 
(ii) the same MLLM that builds the memory also diagnoses it, without a stronger model or gold answer in the loop, and each of its claims must be grounded in the stored memory; 
(iii) it repairs the memory content itself rather than the prompts, after each answer and without knowing whether it was right; 
(iv) beyond attributing observed failures, we inject known errors and replay against matched controls to measure how an error propagates through later operations and what it does to downstream stages.

\section{Problem Formulation}
\label{sec:problem_formulation}

\paragraph{Memory pipeline.}
A multi-turn conversation is a sequence of turns $x_1,\ldots,x_T$.
Turn $x_t=(u_t,I_t,\delta_t)$ consists of the conversation text $u_t$ of that turn, the set $I_t$ of images the user shows in that turn, and the date $\delta_t$ on which the turn takes place.
Memory stores information about the user and the conversation that later questions may need.
The memory $\mathcal{M}_t$ after turn $t$ is a set of \emph{records}.
A record $m$ holds a text body, the images from $I_t$ it keeps, and its source turns $\mathrm{src}(m)$; it is versioned, $m^{(0)},\ldots,m^{(J)}$, where $m^{(0)}$ is the text written from a turn and each later version is the result of a merge with a newer record.
Each turn passes through two stages (left) and each question through two (right):
\begin{equation}
\begin{aligned}
w_t           &= W(x_t),                   &\quad C_q      &= R(q,\ \mathcal{M}_T),\\
\mathcal{M}_t &= U(\mathcal{M}_{t-1},\ w_t), &\quad \hat a_q &= A(q,\ C_q),
\end{aligned}
\label{eq:pipeline}
\end{equation}
\stage{Write} $W$ turns $x_t$ into a new record $w_t$; \stage{Update} $U$ adds $w_t$ to memory, merging it into an existing record when the two describe the same thing; \stage{Retrieve} $R$ selects for a question $q$ (text, possibly with a query image) the records $C_q$ shown to the answerer, with their kept images; \stage{Answer} $A$ produces the answer $\hat a_q$ from $q$ and $C_q$.
Apart from scoring records during retrieval, every stage is a call to the same MLLM, and a log keeps the input and output of every call, including all record versions.

\paragraph{Where an error is.}
Let $a^\star_q$ be the gold answer of $q$ and $F_q$ the facts needed to answer it.
A fact $f\in F_q$ stated at turn $x_t$ reaches the answer along a provenance chain
\begin{equation}
x_t \xrightarrow{\;W\;} m^{(0)} \xrightarrow{\;U\;} \cdots \xrightarrow{\;U\;} m^{(J)}
    \xrightarrow{\;R\;} C_q \xrightarrow{\;A\;} \hat a_q ,
\label{eq:chain}
\end{equation}
where $m$ is the record that holds the content of $x_t$.
We write $z\models f$ when $z$ (a turn, a record version, or a set of records with their images) states $f$ correctly.
A stage is \emph{faulty} for $q$ when its input states a needed fact and its output does not (Table~\ref{tab:taxonomy}); the \emph{source stage} $s^\star_q$ is the earliest faulty stage over the chains of all facts in $F_q$, and $s^\star_q=\textsc{none}$ when $\hat a_q=a^\star_q$.
The stage says \emph{where} a chain breaks and the error type $\varepsilon\in\mathcal{E}_s$ says \emph{how}; an error is therefore a triple $(s,\varepsilon,m^\star_q)$, with $m^\star_q$ the record whose version breaks the chain (none for answer errors).
Table~\ref{tab:taxonomy} lists the error types of each stage. Two arise only because memory is multimodal: \emph{visual loss}, where a fact visible only in the image never reaches the text record, and \emph{visual misdescription}, where the image is described as something similar but different. In \emph{retrieval distraction} the evidence is retrieved but other retrieved records derail the answer, and in \emph{false recall} a question the conversation never addresses is answered with invented content.
A merge can also keep $f$ but bury it in a longer record that retrieval no longer finds; since splitting the merge back restores the answer, we count this as an update loss as well.

\begin{table}[t]
\centering
\footnotesize
\setlength{\tabcolsep}{3pt}
\renewcommand{\arraystretch}{1.15}
\begin{tabularx}{\columnwidth}{@{}l>{\raggedright\arraybackslash}p{1.95cm}>{\raggedright\arraybackslash}X>{\raggedright\arraybackslash}p{1.75cm}@{}}
\toprule
Stage $s$ & Faulty when & Error types $\mathcal{E}_s$ & Repair $\rho_s$ \\
\midrule
Write    & $x_t\models f$\newline $m^{(0)}\not\models f$ & omission, distortion, visual loss, visual misdescription & rewrite from $\mathrm{src}(m)$ \\
Update   & $m^{(j-1)}\models f$\newline $m^{(j)}\not\models f$ or buries $f$ & update loss & re-merge from versions and $\mathrm{src}(m)$ \\
Retrieve & $m^{(J)}\models f$\newline $m\notin C_q$ or $C_q$ derails & retrieval miss, retrieval distraction & re-query \\
Answer   & $C_q\models F_q$\newline $\hat a_q\neq a^\star_q$ & misreasoning, false recall & re-answer from evidence \\
None     & $\hat a_q=a^\star_q$ & --- & no change \\
\bottomrule
\end{tabularx}
\caption{Stages, error types and repairs. A stage is faulty when its input states a needed fact $f$ and its output does not.}
\label{tab:taxonomy}
\end{table}

\paragraph{Self-diagnosis and repair.}
After the system answers $q$, the diagnoser $g$ receives the question, the answer $\hat a_q$ and a view $V_q$ of the log: the retrieved records with their version histories and kept images, and the lower-ranked candidates.
It outputs
\begin{equation}
g(q,\hat a_q,V_q) = (\hat s,\ \hat m,\ \omega),
\label{eq:diag}
\end{equation}
a stage $\hat s\in\{W,U,R,A,\textsc{none}\}$, a target record $\hat m$, and a payload $\omega$ taken from $V_q$: the located evidence for an answer error, a new query for a retrieval error, the lost content for an update error, and for a write error the faulty record itself.
The repair operator $\rho_{\hat s}$ (Table~\ref{tab:taxonomy}) then acts on $\hat m$ with $\omega$; a write repair rewrites $\hat m$ from its source turns $\mathrm{src}(\hat m)$, the only step that reads the original conversation.
Re-retrieving and re-answering on the repaired memory $\mathcal{M}'$ gives a new answer $\hat a'_q$, and the outcome is carried into the next round until the diagnoser reports \textsc{none} or a round limit is reached.
Three constraints define the setting.
(i)~\emph{No external signal}: $g$ sees neither $a^\star_q$, nor the evidence turns, nor whether $\hat a_q$ is correct, so a correct answer should yield $\hat s=\textsc{none}$ and leave the memory unchanged.
(ii)~\emph{Self}: $g$ is the same MLLM that built the memory.
(iii)~\emph{Grounded}: every output must be supported by content that can be located in $V_q$---in a record, in an earlier version of it, or in a kept image---and a repair is kept only if the located evidence supports the new answer.

\paragraph{What we measure.}
Three quantities are reported, each on a fixed question set.
\noindent\textbf{Propagation.} At one operation, an injected error of type $\varepsilon$ is compared with a neutral edit at the same place that preserves every fact; with $\hat a_q^{\varepsilon}$ and $\hat a_q^{n}$ the resulting answers and $c(\cdot)\in\{0,1\}$ correctness, the excess failure rate over the questions $\mathcal{Q}_0$ answered correctly before the edit is
\begin{equation}
P^{\mathrm{excess}}_{\varepsilon}=\frac{1}{|\mathcal{Q}_0|}\sum_{q\in\mathcal{Q}_0}\big[c(\hat a_q^{n})-c(\hat a_q^{\varepsilon})\big].
\label{eq:propagation}
\end{equation}
\noindent\textbf{Localization.} A diagnosis is correct when it names the source stage, $\hat s=s^\star_q$.
For an injected error $s^\star_q$ is the injected stage, known by construction; for a natural failure it follows the definition above and can only be determined with the full trajectory, the gold answer and the source turns; a correctly answered question has $s^\star_q=\textsc{none}$.
Neither label is available to $g$.
\noindent\textbf{Repair.} On a set $\mathcal{Q}$ that contains both initially correct and incorrect answers, with $c_q$ and $c'_q$ correctness before and after repair, the net gain is
\begin{equation}
G_{\mathrm{repair}}=\frac{1}{|\mathcal{Q}|}\sum_{q\in\mathcal{Q}}(c'_q-c_q)=\frac{N_{0\to1}-N_{1\to0}}{|\mathcal{Q}|},
\label{eq:repair}
\end{equation}
where $N_{1\to0}$ counts answers that repair damaged; evaluating only known failures would hide this term.

\providecommand{\pilot}[1]{\ifdefined\textcolor\textcolor{red}{#1}\else#1\fi}
\providecommand{\revise}[1]{\ifdefined\textcolor\textcolor{red}{#1}\else#1\fi}
\providecommand{\emptycell}{\textemdash}
\providecommand{\toprule}{\hline}
\providecommand{\midrule}{\hline}
\providecommand{\bottomrule}{\hline}
\ifdefined\cmidrule\else\def\cmidrule(#1)#2{\cline{#2}}\fi

\section{The \method{} Framework}
\label{sec:method}

\begin{figure*}[t]
  \centering
  \includegraphics[width=\textwidth]{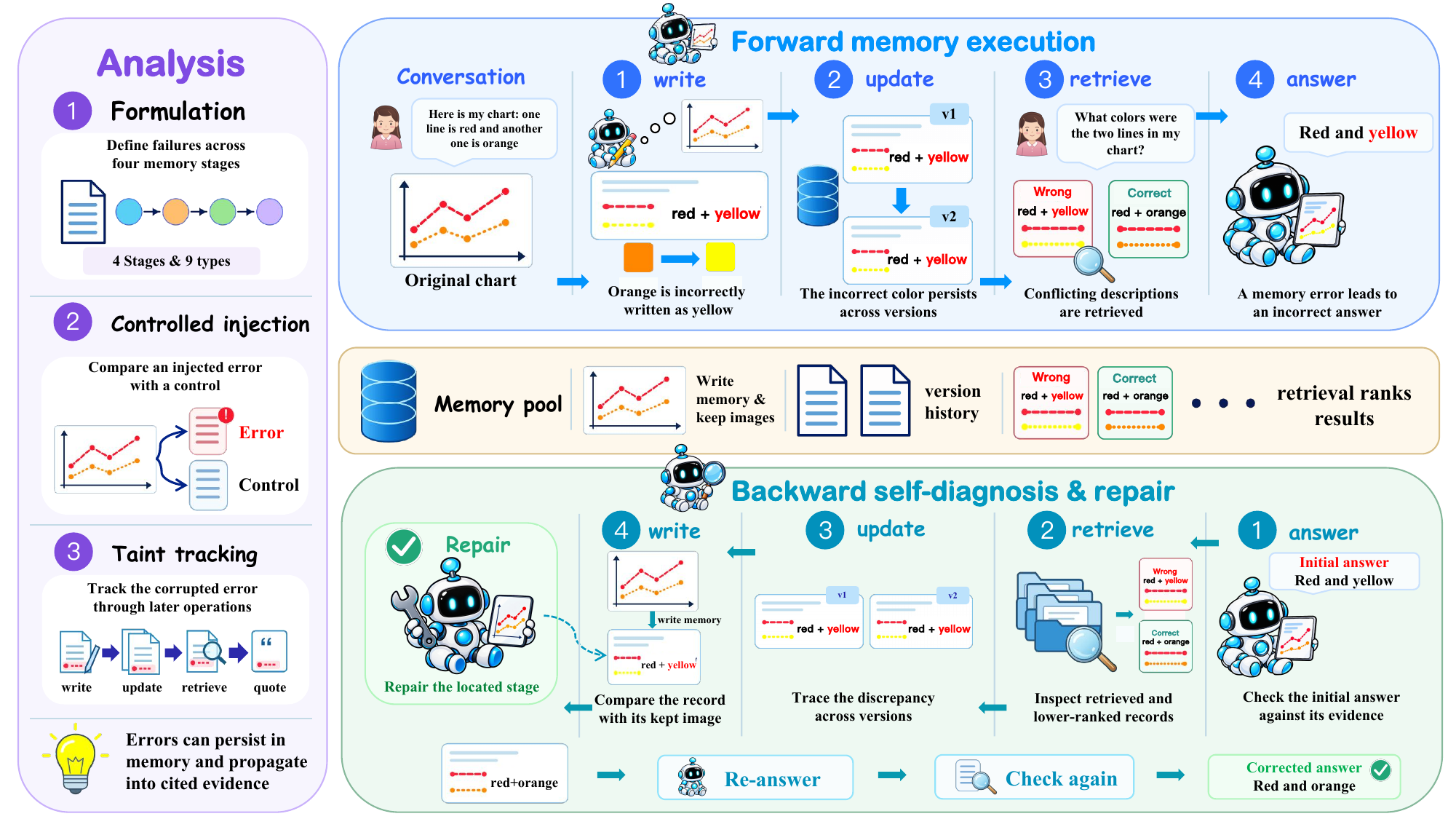}
  \caption{Overview of \method{}. Left: the offline propagation analysis (formulation, controlled injection, taint tracking). Right: the memory system runs forward (top) and logs a memory trace (middle); self-diagnosis checks the stages backward from the answer and repairs the stage it locates (bottom), without the gold answer.}
  \label{fig:framework}
\end{figure*}

\method{} has two parts built on the formulation of Sec.~\ref{sec:problem_formulation}.
The error propagation analysis (Sec.~\ref{sec:propagation}) is an offline study: it plants known errors at one stage and measures what each does downstream, which shows that a wrong record can reach the answer looking like valid evidence (Figure~\ref{fig:framework}, left).
Self-diagnosis and repair (Sec.~\ref{sec:memdoctor}) runs inside the memory system at test time, with only the memory trace and no gold answer, and for this reason checks the stages from the answer backward instead of trusting the evidence an answer cites (Figure~\ref{fig:framework}, right).

\subsection{Error Propagation Analysis}
\label{sec:propagation}

Section~\ref{sec:problem_formulation} defines where an error arises. To see what an error does after it arises, we borrow
the idea of dynamic taint analysis~\citep{newsome2005taintcheck}: we plant a known error at one stage, mark the memory
content derived from it, and follow the mark to the answer.

\paragraph{Controlled injection.}
Each injection targets one \emph{target question} $q$ that the clean system answers correctly and whose evidence lies
in at most three dialogue turns (its \emph{evidence turns}). One stage is corrupted at a time. At the write stage, the
records of the evidence turns either no longer state the fact that $q$ needs (\emph{omission}) or state a plausible
wrong value $\tilde a$ (\emph{distortion}). For image content, the description and the image label are removed and the
image is no longer attached (\emph{visual loss}), or they name something else of the same kind while the image stays
attached (\emph{visual misdescription}). At the update stage, a later merge into an evidence record drops information that
its previous version held (\emph{update loss}); we keep only losses that the merge guard of our system does not catch,
since the guard turns the others into appends. At the retrieval stage, the evidence records are demoted below the retrieval cutoff, so that they are no longer passed to the answerer, without changing the memory (\emph{retrieval miss}),
or the evidence stays in the context while the two records most similar to the question from below the cutoff are lifted above it (\emph{retrieval distraction}); 
its control lifts the two least similar records instead. These seven types cover every memory-side error of Table~\ref{tab:taxonomy}; misreasoning and false recall arise from how the answerer uses intact evidence rather than from memory, and are studied on natural failures only. Qwen3-VL-30B-A3B-Instruct rewrites the evidence records to
introduce the error. If the rewrite of an omission or an update loss is rejected twice, we fall
back to a simple rule: every sentence of the record that contains the answer is deleted. The system model then verifies each edit before replay: the clean evidence records answer $q$, the edited ones do not (judged on the text alone for omission and visual misdescription,
so that the attached image does not hide the error), a distortion makes them yield $\tilde a$, and the other questions
on the same turns remain answerable. We keep edits whose question other records could still answer and flag them as
\emph{redundant}. Except for retrieval errors, the corrupted memory is carried through every later merge and
retrieval operation to the end of the dialogue, and the affected questions are answered again.

\paragraph{Taint tracking.}
The injected edit is the taint source. A record is \emph{tainted} if its content derives from an edited turn: the
record written at that turn and every record it is later merged into. For each target question we check four points
along the path from memory to answer: the error is still present in the tainted records at the end of the dialogue
(\emph{in memory}), a tainted record enters the retrieved context (\emph{retrieved}), the answer quotes a tainted record
as evidence (\emph{quoted}), and the answer is wrong. Unlike program variables, records are rewritten rather than
copied, so a later merge can restore the correct fact. We therefore test the first point by content, asking the system
model the target question from the tainted records alone, instead of only propagating the mark.

\paragraph{Separating the error from replay noise.}
Replaying the dialogue on a modified memory changes later model outputs even when no fact changes, because every later
call sees a different input and decoding is stochastic. We therefore run every injection twice: once
with the error, and once with a \emph{control} edit that reorders the sentences of the same records without changing
any fact. Only the difference between the two runs is attributed to the error. For each answer that becomes wrong, we
also restore just the injected fact in the final memory and ask again, and re-ask the clean and corrupted memories to
catch chance flips: 209 of the 267 wrong answers return to correct, 55 are chance flips or also occur under the
control, and 3 stay wrong after the fix.

\subsection{Self-Diagnosis and Repair}
\label{sec:memdoctor}

Self-diagnosis is the diagnoser $g$ of Sec.~\ref{sec:problem_formulation}, run by the same MLLM that built the memory (Figure~\ref{fig:framework}).
The fault conditions of Table~\ref{tab:taxonomy} are stated in terms of the facts a question needs, which are known only with the gold answer.
\method{} replaces them with the facts the model itself says the question needs, and accepts a statement about where such a fact is only if the fact can be found there in memory.
Since it needs only the memory trace, it can be added to any memory system that logs record versions and retrieval ranks.

\paragraph{Locating the error, from the answer backward.}
The model first breaks $q$ into the facts it needs (at most four) and checks the stages in the reverse order of Eq.~\eqref{eq:chain}, moving to an earlier stage only when the later one is cleared.
\emph{Answer.} For each fact the model names the shown record that states it, and the fact counts as confirmed only if that record or its kept image indeed states it.
If every fact is confirmed and the facts give the system's own answer, the diagnosis is \textsc{none}; if they are confirmed but give another answer, the error lies in \stage{Answer}.
\emph{Retrieve.} Facts missing from the shown records, or located only in records the model itself marks as uncertain, are searched for below the retrieval cutoff (ranks $k{+}1$ to 50); a fact found there points to \stage{Retrieve}.
\emph{Update.} Facts still missing are searched for in the earlier versions of the records; a fact held by an earlier version but gone from the current one points to \stage{Update}.
\emph{Write.} A fact that the records state wrongly in every version, or that no record ever stored, points to \stage{Write}: the error was made when the record was written.
The repair of the stage reached (Table~\ref{tab:taxonomy}) is applied to the record involved, and the system answers again.

\paragraph{Reflection across rounds.}
A repair that does not settle the question should not be repeated.
\method{} keeps a case file of every round: what was located for each fact, which repair was applied, what it changed and the answer it produced.
In each round the model reads the whole file, names why the last attempt failed (wrong stage, wrong record, the repair failed, a fact was not found, or the facts were misread) and proposes one next step.
The proposal is taken only if memory supports it---a proposed retrieval, for instance, must surface a record not shown before that states a needed fact---and otherwise the backward order above is followed, over multiple rounds up to a fixed limit.
Unlike Reflexion~\citep{shinn2023reflexion} and Self-Refine~\citep{madaan2023selfrefine}, which turn task feedback or the model's own critique into a revised output, this reflection needs no feedback and acts on the memory rather than the answer: it reasons over where the error was located and what each repair changed.

\paragraph{Without and with feedback.}
Without feedback, correct and wrong answers go through the same loop, so a repair can also break a correct answer; a new answer therefore replaces the old one only if the located facts support it and it neither gives up nor says less than the old one.
With feedback, as in Reflexion~\citep{shinn2023reflexion}, constraint~(i) of Sec.~\ref{sec:problem_formulation} is relaxed: the model also receives an outcome signal, whether its answer was judged correct. We use this setting only to locate injected errors (Sec.~\ref{sec:propagation_results}).

\section{Experiments}
\label{sec:experiments}
\setcounter{dbltopnumber}{1}
\providecommand{\dup}[1]{{\tiny\textcolor[RGB]{0,128,60}{+#1}}}
\providecommand{\ddown}[1]{{\tiny\textcolor[RGB]{190,30,30}{$-$#1}}}
\providecommand{\dzero}{{\tiny\textcolor{gray}{0.0}}}
\providecommand{\hd}[1]{\makebox[\mgnumw][c]{#1}}
\begin{table*}[t]
\centering
\small
\setlength{\tabcolsep}{0pt}
\renewcommand{\arraystretch}{1.12}
\ifdefined\mgnumw\else\newlength{\mgnumw}\fi\settowidth{\mgnumw}{00.00}
\begin{tabular*}{\textwidth}{@{\extracolsep{\fill}}l@{\hspace{8pt}}*{9}{r@{\hspace{15pt}}}r@{\hspace{15pt}}}
\toprule
Method & \hd{FR} & \hd{MR} & \hd{TR} & \hd{VR} & \hd{TTL} & \hd{VS} & \hd{CD} & \hd{KR} & \hd{AR} & \hd{\textbf{Overall}} \\
\midrule
\multicolumn{11}{@{}l}{\textbf{Qwen3-VL-8B-Instruct Backbone}} \\
\addlinespace[2pt]
TextRAG & 70.78 & 74.27 & 66.67 & 41.95 & \underline{75.67} & 56.54 & \underline{66.67} & 64.20 & 94.57 & 68.44 \\
MuRAG & 71.69 & \underline{78.16} & 69.92 & 49.43 & \textbf{85.76} & \textbf{75.82} & 62.96 & 62.96 & 94.02 & \underline{75.16} \\
A-Mem & 68.49 & 61.17 & 69.92 & 55.17 & 64.69 & 67.97 & 32.10 & 67.90 & \underline{96.74} & 66.80 \\
Mem0 & 70.32 & 58.74 & 33.33 & 40.80 & 67.66 & 8.50 & 33.33 & 54.32 & \textbf{97.28} & 52.07 \\
Omni-SimpleMem & 71.23 & 62.14 & 71.54 & 57.47 & 69.44 & 68.95 & 34.57 & 61.73 & 93.48 & 68.21 \\
V-Mem & 79.91 & \textbf{83.01} & 68.29 & \underline{60.34} & 47.48 & 68.30 & 55.56 & 66.67 & 70.65 & 66.22 \\
\midrule
\multicolumn{11}{@{}l}{\textbf{Qwen3-VL-4B-Instruct Backbone}} \\
\addlinespace[2pt]
Ours w/o repair & \textbf{80.82} & \underline{78.16} & \underline{76.42} & 54.02 & 73.59 & 66.01 & 45.68 & \textbf{74.07} & 92.93 & 72.71 \\
\textbf{Ours} & \underline{80.37}\rlap{\hspace{1.5pt}\ddown{0.5}} & \underline{78.16}\rlap{\hspace{1.5pt}\dzero} & \textbf{80.49}\rlap{\hspace{1.5pt}\dup{4.1}} & \textbf{62.07}\rlap{\hspace{1.5pt}\dup{8.1}} & 75.37\rlap{\hspace{1.5pt}\dup{1.8}} & \underline{70.59}\rlap{\hspace{1.5pt}\dup{4.6}} & \textbf{72.84}\rlap{\hspace{1.5pt}\dup{27.2}} & \underline{71.60}\rlap{\hspace{1.5pt}\ddown{2.5}} & 88.59\rlap{\hspace{1.5pt}\ddown{4.3}} & \textbf{75.63}\rlap{\hspace{1.5pt}\dup{2.9}} \\
\bottomrule
\end{tabular*}
\caption{\textbf{Mem-Gallery accuracy (\%) by task.} All methods cover 20 scenes. Overall is question-weighted. Small numbers next to Ours give the change from Ours w/o repair, in points. \textbf{Bold} and \underline{underline} mark the highest and second-highest reported accuracy in each column, respectively.}\label{tab:memgallery_main}
\vspace{2pt}
\parbox{\textwidth}{\footnotesize FR: factual retrieval; MR: multi-entity reasoning; TR: temporal reasoning; VR: visual-centric reasoning; TTL: test-time learning; VS: visual-centric search; CD: conflict detection; KR: knowledge resolution; AR: answer refusal.}
\end{table*}

\begin{table*}[t]
\centering
\small
\setlength{\tabcolsep}{5pt}
\begin{tabular}{llrccccc}
\toprule
 & & & \multicolumn{3}{c}{Tainted record} & & \\
\cmidrule(lr){4-6}
Stage & Error & $N$ & in memory & retrieved & quoted & Wrong answer (inj.\ / ctrl.) & $\Delta$ [95\% CI] \\
\midrule
Write    & Omission              & 130 & 65\%  & 85\%  & 62\%  & 28\% / 3\%   & +25 [+19, +34] \\
Write    & Distortion            & 133 & 80\%  & 100\% & 97\%  & 45\% / 2\%   & +43 [+35, +53] \\
Write    & Visual loss           & 12  & 50\%  & 100\% & 67\%  & 50\% / 0\%   & +50 [+25, +75] \\
Write    & Visual misdescription & 12  & 17\%  & 100\% & 0\%   & 33\% / 0\%   & +33 [0, +59] \\
Update   & Update loss           & 11  & 73\%  & 91\%  & 80\%  & 45\% / 0\%   & +45 [+17, +80] \\
Retrieve & Retrieval miss        & 289 & --   & 0\%   & --   & 45\% / --   & -- \\
Retrieve & Retrieval distraction & 320 & --   & 100\% & 8\%   & 8\% / 8\%    & 0 [$-$4, +4] \\
\bottomrule
\end{tabular}
\caption{\textbf{Following injected errors from memory to answer} (Mem-Gallery, 20 scenes, Qwen3-VL-30B-A3B).
$N$: accepted injections. \emph{In memory} / \emph{retrieved}: the tainted record is still present at the end of the
dialogue / among the top-8 records; \emph{quoted}: share of the wrong answers that quote it. \emph{Wrong answer}: after
the injection / after the control edit. $\Delta$: paired difference in points (95\% CI over scenes).
Retrieval errors change no memory; a retrieval miss has no control.}
\label{tab:propagation}
\end{table*}

\subsection{Setup}
\label{sec:setup}

\paragraph{Benchmarks.}
Mem-Gallery~\citep{bei2026memgallery} consists of multimodal multi-session conversations, each a scene with its own question set.
MemLens~\citep{ren2026memlens} gives each question its own history, so memory is built once per question; we use its 195-question agent subset at the 32K length.

\paragraph{Models.}
We use the 4B, 8B, and 30B-A3B Instruct variants of Qwen3-VL~\citep{bai2025qwen3vl}.
For the comparison on all 20 Mem-Gallery scenes (1{,}711 questions) and MemLens (195 questions), \method{} uses Qwen3-VL-4B-Instruct for memory construction, answering, diagnosis, and repair, while the six external baselines use Qwen3-VL-8B-Instruct.
On Mem-Gallery, answers are scored using its category-specific rules and graded rubric, with Qwen3-VL-30B-A3B-Instruct as the LLM judge.
A score of at least 0.75 counts as correct.
On MemLens, the same judge assigns binary correctness scores using its official task-specific rubrics.
On both benchmarks, \method{} runs at most three rounds per question without feedback, the judge only scores final answers and never informs the repair loop, and we report the answer of the last round. Repairs made for one question do not carry over to the others.
The propagation analysis (Table~\ref{tab:propagation}) uses Qwen3-VL-30B-A3B-Instruct as the memory system, the editor that injects errors and the judge.

\paragraph{Memory system.}
A turn is written into one record, and \stage{Update} compares it with the three most similar records, merging it into one of them or adding it.
\stage{Retrieve} embeds the question with BGE-M3~\citep{chen2024m3}, describes a query image and adds the description to the query, and passes up to six records, with at most four of their images, to \stage{Answer}.
When the answer cites no evidence that can be located in these records, the next records are added and the question is answered again.

\paragraph{Baselines.}
We include two retrieval baselines and four memory systems, all using Qwen3-VL-8B-Instruct as the answering backbone.
\textbf{TextRAG} is our text-only retrieval baseline following the RAG paradigm~\citep{lewis2020rag}; it indexes raw conversation turns with BGE-M3.
Our adapted \textbf{MuRAG} baseline~\citep{chen2022murag} instead uses GME-Qwen2-VL-2B-Instruct~\citep{zhang2025gme} to retrieve from fused text--image representations, including a question image when available.
On Mem-Gallery, both retrieve eight turns, supply at most four retrieved images to the answerer, and make a single answering call without deeper retrieval.
\textbf{A-Mem}~\citep{xu2025amem}, \textbf{Mem0}~\citep{chhikara2025mem0}, and \textbf{Omni-SimpleMem}~\citep{liu2026omnisimplemem} construct persistent memories before retrieval and answering.
\textbf{V-Mem}~\citep{kang2026vmem} is an additional multimodal memory baseline, evaluated without our diagnosis or repair loop.
On MemLens, the baselines retrieve ten turns or memory records, except for V-Mem, which uses its native budget of twelve.
We also report the initial answers of our own 4B memory system before repair (\textbf{Ours w/o repair}).

\paragraph{Source-stage labels.}
For injected errors the source stage is known by construction (Sec.~\ref{sec:propagation}); we evaluate localization on these errors with the 30B system of the propagation analysis.

\paragraph{Metrics.}
We report accuracy, and on Mem-Gallery also the mean judge score; localization accuracy against the source stage (Sec.~\ref{sec:problem_formulation}); and the net repair gain of Eq.~\eqref{eq:repair} together with the number of answers that repair breaks.

\subsection{Memory Question Answering}
\label{sec:main_results}

\paragraph{Main results.}
Table~\ref{tab:memgallery_main} reports accuracy on Mem-Gallery.
Before repair, our memory system on a 4B backbone reaches 72.71\%, above every 8B baseline except MuRAG (75.16\%).
With self-diagnosis and repair on every answer, \method{} raises accuracy to 75.63\% (+2.92 points; mean judge score 0.740 to 0.771), the highest among all systems despite the smaller backbone.
It recovers 94 of the 467 initial failures (20.1\%) and breaks 44 of the 1{,}244 correct answers (3.5\%), a net gain of 50 questions.
The gains are largest on conflict detection (+27.2 points), visual-centric reasoning (+8.1), visual-centric search (+4.6) and temporal reasoning (+4.1), while answer refusal ($-4.3$), knowledge resolution ($-2.5$) and factual retrieval ($-0.5$) lose slightly.

\paragraph{Findings.}
\textbf{Repairing the located stage fixes what re-reading cannot.}
The baselines answer once from the retrieved records, so a fact ranked below the cutoff or stored wrongly stays out of reach; \method{} repairs the stage where the fact is.
\textbf{Conflict and visual questions gain most.}
Conflict questions often hinge on a detail lost when the turn was written, which only the write repair can restore; memory systems that store rewritten text have no such path and trail on this task (32--35\%).
Visual questions gain from searching below the retrieval cutoff (Sec.~\ref{sec:memdoctor}), which brings up image records just below it.

\paragraph{MemLens.}
\providecommand{\mldup}[1]{{\fontsize{5pt}{6pt}\selectfont\textcolor[RGB]{0,128,60}{+#1}}}
\providecommand{\mlddown}[1]{{\fontsize{5pt}{6pt}\selectfont\textcolor[RGB]{190,30,30}{$-$#1}}}
\providecommand{\mldzero}{{\fontsize{5pt}{6pt}\selectfont\textcolor{gray}{0.0}}}
\providecommand{\mld}[1]{\hspace{0.5pt}\raisebox{-0.5ex}{#1}}
\begin{table}[t]
\centering
\fontsize{7.5pt}{9pt}\selectfont
\setlength{\tabcolsep}{0pt}
\renewcommand{\arraystretch}{1.08}
\begin{tabular*}{\columnwidth}{@{\extracolsep{\fill}}l@{\hspace{3pt}}*{5}{r@{}l@{\hspace{1.5pt}}}r@{}l}
\toprule
Method & IE & & MSR & & TR & & KU & & AR & & \textbf{Overall} & \\
\midrule
\multicolumn{13}{@{}l}{\textbf{Qwen3-VL-8B-Instruct Backbone}} \\
\addlinespace[1pt]
TextRAG & 42.6 & & 17.1 & & 50.0 & & \underline{34.5} & & 100.0 & & 45.1 & \\
MuRAG & 37.7 & & 17.1 & & 52.1 & & 27.6 & & 100.0 & & 43.1 & \\
A-Mem & 42.6 & & 11.4 & & 47.9 & & 31.0 & & 100.0 & & 43.1 & \\
Mem0 & 18.0 & & \underline{22.9} & & 39.6 & & \underline{34.5} & & 100.0 & & 35.9 & \\
Omni-SimpleMem & 37.7 & & 14.3 & & 43.8 & & 27.6 & & 100.0 & & 40.5 & \\
V-Mem & 50.8 & & 14.3 & & 45.8 & & \textbf{37.9} & & 100.0 & & 46.7 & \\
\midrule
\multicolumn{13}{@{}l}{\textbf{Qwen3-VL-4B-Instruct Backbone}} \\
\addlinespace[1pt]
Ours w/o repair & \underline{54.1} & & \underline{22.9} & & \underline{58.3} & & \underline{34.5} & & 100.0 & & \underline{51.8} & \\
\textbf{Ours} & \textbf{60.7} & \mld{\mldup{6.6}} & \textbf{25.7} & \mld{\mldup{2.9}} & \textbf{62.5} & \mld{\mldup{4.2}} & \underline{34.5} & \mld{\mldzero} & 100.0 & \mld{\mldzero} & \textbf{55.4} & \mld{\mldup{3.6}} \\
\bottomrule
\end{tabular*}
\caption{\textbf{MemLens accuracy (\%) by task} (32K, 195-question agent subset). Overall is question-weighted. Small numbers next to Ours give the change from Ours w/o repair, in points. \textbf{Bold} and \underline{underline} mark the highest and second-highest accuracy in each column. IE: information extraction; MSR: multi-session reasoning; TR: temporal reasoning; KU: knowledge update; AR: answer refusal.}
\label{tab:memlens_main}
\end{table}

Table~\ref{tab:memlens_main} repeats the comparison on MemLens, where every question has its own history.
Before repair, our 4B memory system reaches 51.8\%, above every 8B baseline (the strongest, V-Mem, reaches 46.7\%).
With self-diagnosis and repair, \method{} reaches 55.4\% (+3.6 points), 8.7 points above V-Mem.
It recovers 10 of the 94 initial failures (10.6\%) and breaks 3 of the 101 correct answers (3.0\%).
The gains come from information extraction (+6.6 points), temporal reasoning (+4.2) and multi-session reasoning (+2.9); knowledge update is unchanged, and every system answers all refusal questions correctly.
\textbf{A total is diagnosed as one missing fact.} Multi-session reasoning asks for a sum or a count over several sessions. In 81\% of its failures, the diagnoser lists the total itself as the fact the answer needs, rather than the items that make it up; no record states the total, so most of these diagnoses (18 of 22) blame \stage{Write}.

\subsection{How Errors Propagate}
\label{sec:propagation_results}

Table~\ref{tab:propagation} shows that \textbf{wrong values can persist through memory updates and reach the answer as apparently valid evidence}.
Distortions increase the failure rate by 43 points over the control, and the tainted record is cited in 97\% of the wrong answers.
By the end of the dialogue, 80\% of distortions remain in memory, compared with 65\% of omissions.
Update loss and visual loss also affect downstream answering, with observed failure rates of 45\% and 50\%, respectively, compared with 0\% in their corresponding controls.

\textbf{The impact depends on whether alternative evidence is available.}
As shown in Appendix~\ref{app:propagation} (Table~\ref{tab:redundancy}), distortions increase the failure rate over the control by 68 percentage points when no alternative evidence is available, compared with 10 points when other records or retained images still support the answer.
Retrieval distraction produces no observed increase in failures (8\% under both conditions, over 320 injections).
Visual and update errors have fewer accepted injections, with only 11--12 cases per type.

On the 247 injected failures, self-diagnosis with outcome feedback and up to five rounds locates the source stage in 67\% of cases and recovers the correct answer in 62\% (Appendix~\ref{app:feedback}).
Retrieval misses reach 71\% localization and 69\% recovery.
\textbf{Distortions remain difficult to diagnose:} without feedback, the model finds nothing wrong in two thirds of these cases; with feedback, it mostly blames the answering stage because the cited record appears to support the answer.

\subsection{Plug-in Evaluation}
\label{sec:plugin_results}

Agent memory systems differ in how they store, update, and retrieve information, but errors within these pipelines can compromise downstream answers.
\method{} addresses this shared reliability problem by tracing failures to their source stages and applying targeted repairs.
To evaluate whether this capability can serve as a reusable repair layer, we attach \method{} to MuRAG and A-Mem.
Both backends and the repair loop use Qwen3-VL-8B-Instruct.
On MemLens, \method{} improves their accuracies by 7.69 and 4.10 percentage points over their respective paired baselines.
On Mem-Gallery, the corresponding gains are 1.46 and 1.52 percentage points.
Detailed experimental settings and results are provided in Appendix~\ref{app:backend_repair}.

\section{Conclusion}
\label{sec:conclusion}

We studied where multimodal agent memory fails and how those failures propagate.
By dividing the pipeline into four stages and replacing one stage's output at a time, we traced most failures to a source stage.
Most of the failures we traced originated before answer generation.
Controlled error injections showed why these errors are hard to detect from the answer alone.
An incorrect value can survive later merges and be cited as evidence.
\method{} uses the model that built the memory to check the stages backward from its answer, ground its diagnosis in stored evidence, and repair the stage it identifies without being told whether the answer was correct.
On a 4B backbone, \method{} raised accuracy from 72.71\% to 75.63\% on Mem-Gallery and from 51.8\% to 55.4\% on MemLens, outperforming all evaluated 8B baselines on both benchmarks.
As a plug-in, \method{} improved MuRAG and A-Mem over their respective baselines by 1.46 and 1.52 percentage points on Mem-Gallery.
The corresponding gains on MemLens were 7.69 and 4.10 points (Appendix~\ref{app:backend_repair}).
Detecting errors that are consistent across memory remains an open problem.
When every record and its earlier versions give the same incorrect value, a check against memory can still accept that value as supported.

\FloatBarrier
\section*{Limitations}

\paragraph{Single and consistent errors.}
Our propagation analysis plants one error at a time, so it shows how a single error travels through memory but not how
several errors interact, for instance a wrong value written into more than one record, or a later merge that combines an
error with other content. Such errors are also the hardest to diagnose: when every record and every version states the
same wrong value and no kept image contradicts it, every check grounded in memory passes, and without feedback the model
finds nothing wrong in two thirds of the injected distortions. Errors that memory states consistently likely need a
signal from outside the memory to be found.

\paragraph{Cost of repairing without feedback.}
Without the gold answer, the model cannot tell whether an answer is right, so it checks every answer and sometimes changes
a correct one:
on Mem-Gallery, 3.5\% of the correct answers are broken, and answer refusal loses 4.3 points as later rounds overturn
correct refusals. Checking every answer also adds model calls to every question.

\paragraph{Repairs within one question.}
In our evaluation a repair made for one question is discarded before the next, so we do not measure whether a repaired
record also helps later questions, or whether repairs accumulate new errors over time.

\bibliography{main}

\begin{thebibliography}{36}
\providecommand{\natexlab}[1]{#1}

\bibitem[{Bai et~al.(2025{\natexlab{a}})Bai, Cai, Chen, Chen, Chen
  et~al.}]{bai2025qwen3vl}
Shuai Bai, Yuxuan Cai, Ruizhe Chen, Keqin Chen, Xionghui Chen, et~al.
  2025{\natexlab{a}}.
\newblock \href {https://arxiv.org/abs/2511.21631} {{Qwen3-VL} technical
  report}.
\newblock \emph{arXiv preprint arXiv:2511.21631}.

\bibitem[{Bai et~al.(2025{\natexlab{b}})Bai, Chen, Liu, Wang, Ge, Song, Dang,
  Wang, Wang, Tang et~al.}]{bai2025qwen25vl}
Shuai Bai, Keqin Chen, Xuejing Liu, Jialin Wang, Wenbin Ge, Sibo Song, Kai
  Dang, Peng Wang, Shijie Wang, Jun Tang, et~al. 2025{\natexlab{b}}.
\newblock {Qwen2.5-VL} technical report.
\newblock \emph{arXiv preprint arXiv:2502.13923}.

\bibitem[{Bei et~al.(2026)Bei, Wei, Ning, Zhao, Liu, Lin, Zhu, Hamann, He, and
  Tong}]{bei2026memgallery}
Yuanchen Bei, Tianxin Wei, Xuying Ning, Yanjun Zhao, Zhining Liu, Xiao Lin,
  Yada Zhu, Hendrik Hamann, Jingrui He, and Hanghang Tong. 2026.
\newblock \href {https://doi.org/10.18653/v1/2026.acl-long.1892}
  {{Mem-Gallery}: Benchmarking multimodal long-term conversational memory for
  {MLLM} agents}.
\newblock In \emph{Proceedings of the 64th Annual Meeting of the Association
  for Computational Linguistics (Volume 1: Long Papers)}, pages 40750--40784.
  Association for Computational Linguistics.

\bibitem[{Bo et~al.(2026)Bo, Zhang, Sun, Wu, Xie, Tan, Chen, He, Li, Zhao,
  Wang, and Li}]{bo2026vilomem}
Weihao Bo, Shan Zhang, Yanpeng Sun, Jingjing Wu, Qunyi Xie, Xiao Tan, Kunbin
  Chen, Wei He, Xiaofan Li, Na~Zhao, Jingdong Wang, and Zechao Li. 2026.
\newblock \href
  {https://openaccess.thecvf.com/content/CVPR2026/html/Bo_ViLoMem_Agentic_Learner_with_Grow-and-Refine_Multimodal_Semantic_Memory_CVPR_2026_paper.html}
  {{ViLoMem}: Agentic learner with grow-and-refine multimodal semantic memory}.
\newblock In \emph{Proceedings of the IEEE/CVF Conference on Computer Vision
  and Pattern Recognition (CVPR)}, pages 5476--5486.

\bibitem[{Chen et~al.(2024)Chen, Xiao, Zhang, Luo, Lian, and Liu}]{chen2024m3}
Jianlyu Chen, Shitao Xiao, Peitian Zhang, Kun Luo, Defu Lian, and Zheng Liu.
  2024.
\newblock \href {https://doi.org/10.18653/v1/2024.findings-acl.137}
  {{M3-Embedding}: Multi-linguality, multi-functionality, multi-granularity
  text embeddings through self-knowledge distillation}.
\newblock In \emph{Findings of the Association for Computational Linguistics:
  ACL 2024}, pages 2318--2335. Association for Computational Linguistics.

\bibitem[{Chen et~al.(2022)Chen, Hu, Chen, Verga, and Cohen}]{chen2022murag}
Wenhu Chen, Hexiang Hu, Xi~Chen, Pat Verga, and William Cohen. 2022.
\newblock \href {https://doi.org/10.18653/v1/2022.emnlp-main.375} {{MuRAG:
  Multimodal Retrieval-Augmented Generator for Open Question Answering over
  Images and Text}}.
\newblock In \emph{Proceedings of the 2022 Conference on Empirical Methods in
  Natural Language Processing}, pages 5558--5570. Association for Computational
  Linguistics.

\bibitem[{Chhikara et~al.(2025)Chhikara, Khant, Aryan, Singh, and
  Yadav}]{chhikara2025mem0}
Prateek Chhikara, Dev Khant, Saket Aryan, Taranjeet Singh, and Deshraj Yadav.
  2025.
\newblock \href {https://arxiv.org/abs/2504.19413} {{Mem0}: Building
  production-ready {AI} agents with scalable long-term memory}.
\newblock \emph{arXiv preprint arXiv:2504.19413}.

\bibitem[{Deng et~al.(2026)Deng, Zhong, Peng, Lu, Wu, Li, Xu, Yao, Fang, Cao,
  Guo, Yuan, Ma, Yu, Hu, Dong, Zhu, and Zhang}]{deng2026memtrace}
Xinle Deng, Ruobin Zhong, Hujin Peng, Xiaoben Lu, Yanzhe Wu, Guang Li, Buqiang
  Xu, Yunzhi Yao, Jizhan Fang, Haoliang Cao, Junjie Guo, Yuan Yuan, Ziqing Ma,
  Yuanqiang Yu, Rui Hu, Baohua Dong, Hangcheng Zhu, and Ningyu Zhang. 2026.
\newblock \href {https://arxiv.org/abs/2605.28732} {{MemTrace}: Tracing and
  attributing errors in large language model memory systems}.
\newblock \emph{arXiv preprint arXiv:2605.28732}.

\bibitem[{Fang et~al.(2026)Fang, Deng, Xu, Jiang, Tang, Xu, Deng, Yao, Wang,
  Qiao, Chen, and Zhang}]{fang2026lightmem}
Jizhan Fang, Xinle Deng, Haoming Xu, Ziyan Jiang, Yuqi Tang, Ziwen Xu, Shumin
  Deng, Yunzhi Yao, Mengru Wang, Shuofei Qiao, Huajun Chen, and Ningyu Zhang.
  2026.
\newblock \href
  {https://proceedings.iclr.cc/paper_files/paper/2026/hash/a05b72653ec5b473732129829ae04195-Abstract-Conference.html}
  {{LightMem}: Lightweight and efficient memory-augmented generation}.
\newblock In \emph{International Conference on Learning Representations
  (ICLR)}.

\bibitem[{Feng et~al.(2026)Feng, Xu, Chen, Dai, Wu, Li, Zeng, Xie, Liang, Lu,
  and Zhang}]{feng2026m2a}
Junyu Feng, Binxiao Xu, Jiayi Chen, Mengyu Dai, Cenyang Wu, Haodong Li, Bohan
  Zeng, Yunliu Xie, Hao Liang, Ming Lu, and Wentao Zhang. 2026.
\newblock \href {https://arxiv.org/abs/2602.07624} {{M2A: Multimodal Memory
  Agent with Dual-Layer Hybrid Memory for Long-Term Personalized
  Interactions}}.
\newblock \emph{arXiv preprint arXiv:2602.07624}.

\bibitem[{Jiang et~al.(2026{\natexlab{a}})Jiang, Li, Li, and
  Li}]{jiang2026magma}
Dongming Jiang, Yi~Li, Guanpeng Li, and Bingzhe Li. 2026{\natexlab{a}}.
\newblock \href {https://doi.org/10.18653/v1/2026.acl-long.1709} {{MAGMA: A
  Multi-Graph based Agentic Memory Architecture for AI Agents}}.
\newblock In \emph{Proceedings of the 64th Annual Meeting of the Association
  for Computational Linguistics (Volume 1: Long Papers)}, pages 36848--36865.
  Association for Computational Linguistics.

\bibitem[{Jiang et~al.(2026{\natexlab{b}})Jiang, Li, Wei, Yang, Kishore, Zhao,
  Kang, Hu, Chen, Li, and Li}]{jiang2026anatomy}
Dongming Jiang, Yi~Li, Songtao Wei, Jinxin Yang, Ayushi Kishore, Alysa Zhao,
  Dingyi Kang, Xu~Hu, Feng Chen, Qiannan Li, and Bingzhe Li.
  2026{\natexlab{b}}.
\newblock \href {https://arxiv.org/abs/2602.19320} {{Anatomy of Agentic Memory:
  Taxonomy and Empirical Analysis of Evaluation and System Limitations}}.
\newblock \emph{arXiv preprint arXiv:2602.19320}.

\bibitem[{Kang et~al.(2026)Kang, Jiang, Li, Li, and Li}]{kang2026vmem}
Dingyi Kang, Dongming Jiang, Yi~Li, Guanpeng Li, and Bingzhe Li. 2026.
\newblock \href {https://arxiv.org/abs/2608.01543} {{V-Mem}: Modality-routed
  retrieval for long-term multimodal agentic memory}.
\newblock \emph{arXiv preprint arXiv:2608.01543}.

\bibitem[{Levy et~al.(2024)Levy, Jacoby, and Goldberg}]{levy2024same}
Mosh Levy, Alon Jacoby, and Yoav Goldberg. 2024.
\newblock \href {https://aclanthology.org/2024.acl-long.818} {Same task, more
  tokens: the impact of input length on the reasoning performance of large
  language models}.
\newblock In \emph{Proceedings of the 62nd Annual Meeting of the Association
  for Computational Linguistics (Volume 1: Long Papers)}, pages 15339--15353.

\bibitem[{Lewis et~al.(2020)Lewis, Perez, Piktus, Petroni, Karpukhin, Goyal,
  K{\"u}ttler, Lewis, Yih, Rockt{\"a}schel, Riedel, and Kiela}]{lewis2020rag}
Patrick Lewis, Ethan Perez, Aleksandra Piktus, Fabio Petroni, Vladimir
  Karpukhin, Naman Goyal, Heinrich K{\"u}ttler, Mike Lewis, Wen-tau Yih, Tim
  Rockt{\"a}schel, Sebastian Riedel, and Douwe Kiela. 2020.
\newblock \href {https://arxiv.org/abs/2005.11401} {Retrieval-augmented
  generation for knowledge-intensive {NLP} tasks}.
\newblock In \emph{Advances in Neural Information Processing Systems}.

\bibitem[{Li et~al.(2025)Li, Song, Wang, Niu, Chen, Yang, Xi, Lai, Zhao, Wang,
  Ren, Lin, Huo, Chen, Chen, Li, Yin, Yu, Tang, Yang, Xu, and
  Xiong}]{li2025memos}
Zhiyu Li, Shichao Song, Hanyu Wang, Simin Niu, Ding Chen, Jiawei Yang, Chenyang
  Xi, Huayi Lai, Jihao Zhao, Yezhaohui Wang, Junpeng Ren, Zehao Lin, Jiahao
  Huo, Tianyi Chen, Kai Chen, Kehang Li, Zhiqiang Yin, Qingchen Yu, Bo~Tang,
  and 3 others. 2025.
\newblock \href {https://arxiv.org/abs/2505.22101} {{MemOS: An Operating System
  for Memory-Augmented Generation (MAG) in Large Language Models}}.
\newblock \emph{arXiv preprint arXiv:2505.22101}.

\bibitem[{Liu et~al.(2026{\natexlab{a}})Liu, Yang, Pu, Liu, Long, Guo, Chen,
  Weng, Kochkina, Kaur, Smiley, Liu, Zou, Liu, Bu, Peng, and
  Wang}]{liu2026worldmemarena}
Chengzhi Liu, Yuzhe Yang, Sophia~Xiao Pu, Yepeng Liu, Lin Long, Yichen Guo, Nuo
  Chen, Zhaotian Weng, Elena Kochkina, Simerjot Kaur, Charese Smiley, Xiaomo
  Liu, James Zou, Sheng Liu, Yuheng Bu, Songyou Peng, and Xin~Eric Wang.
  2026{\natexlab{a}}.
\newblock \href {https://arxiv.org/abs/2605.29341} {{WorldMemArena}: Evaluating
  multimodal agent memory through action--world interaction}.
\newblock \emph{arXiv preprint arXiv:2605.29341}.

\bibitem[{Liu et~al.(2023)Liu, Li, Wu, and Lee}]{liu2023llava}
Haotian Liu, Chunyuan Li, Qingyang Wu, and Yong~Jae Lee. 2023.
\newblock Visual instruction tuning.
\newblock In \emph{Advances in Neural Information Processing Systems
  (NeurIPS)}.

\bibitem[{Liu et~al.(2026{\natexlab{b}})Liu, Ling, Qiu, Liu, Han, Xia, Tu,
  Zheng, Xie, Fleming, Ding, and Yao}]{liu2026omnisimplemem}
Jiaqi Liu, Zipeng Ling, Shi Qiu, Yanqing Liu, Siwei Han, Peng Xia, Haoqin Tu,
  Zeyu Zheng, Cihang Xie, Charles Fleming, Mingyu Ding, and Huaxiu Yao.
  2026{\natexlab{b}}.
\newblock \href {https://arxiv.org/abs/2604.01007} {{Omni-SimpleMem:
  Autoresearch-Guided Discovery of Lifelong Multimodal Agent Memory}}.
\newblock \emph{arXiv preprint arXiv:2604.01007}.

\bibitem[{Liu et~al.(2026{\natexlab{c}})Liu, Su, Xia, Han, Zheng, Xie, Ding,
  and Yao}]{liu2026simplemem}
Jiaqi Liu, Yaofeng Su, Peng Xia, Siwei Han, Zeyu Zheng, Cihang Xie, Mingyu
  Ding, and Huaxiu Yao. 2026{\natexlab{c}}.
\newblock \href {https://arxiv.org/abs/2601.02553} {{SimpleMem: Efficient
  Lifelong Memory for LLM Agents}}.
\newblock \emph{arXiv preprint arXiv:2601.02553}.

\bibitem[{Liu et~al.(2025)Liu, Sun, Cheng, Lei, Chen, Wen, Yang, Fu, Cai, Deng,
  Yu, Hu, Shi, and Wang}]{liu2025memverse}
Junming Liu, Yifei Sun, Weihua Cheng, Haodong Lei, Yirong Chen, Licheng Wen,
  Xuemeng Yang, Daocheng Fu, Pinlong Cai, Nianchen Deng, Yi~Yu, Shuyue Hu,
  Botian Shi, and Ding Wang. 2025.
\newblock \href {https://arxiv.org/abs/2512.03627} {{MemVerse: Multimodal
  Memory for Lifelong Learning Agents}}.
\newblock \emph{arXiv preprint arXiv:2512.03627}.

\bibitem[{Liu et~al.(2024)Liu, Lin, Hewitt, Paranjape, Bevilacqua, Petroni, and
  Liang}]{liu2024lost}
Nelson~F. Liu, Kevin Lin, John Hewitt, Ashwin Paranjape, Michele Bevilacqua,
  Fabio Petroni, and Percy Liang. 2024.
\newblock \href {https://doi.org/10.1162/tacl_a_00638} {Lost in the middle: How
  language models use long contexts}.
\newblock \emph{Transactions of the Association for Computational Linguistics},
  12:157--173.

\bibitem[{Long et~al.(2026)Long, He, Ye, Pan, Lin, Li, Zhao, and
  Li}]{long2026m3agent}
Lin Long, Yichen He, Wentao Ye, Yiyuan Pan, Yuan Lin, Hang Li, Junbo Zhao, and
  Wei Li. 2026.
\newblock \href {https://arxiv.org/abs/2508.09736} {Seeing, listening,
  remembering, and reasoning: A multimodal agent with long-term memory}.
\newblock In \emph{International Conference on Learning Representations
  (ICLR)}.

\bibitem[{Madaan et~al.(2023)Madaan, Tandon, Gupta, Hallinan, Gao, Wiegreffe,
  Alon, Dziri, Prabhumoye, Yang et~al.}]{madaan2023selfrefine}
Aman Madaan, Niket Tandon, Prakhar Gupta, Skyler Hallinan, Luyu Gao, Sarah
  Wiegreffe, Uri Alon, Nouha Dziri, Shrimai Prabhumoye, Yiming Yang, et~al.
  2023.
\newblock Self-refine: Iterative refinement with self-feedback.
\newblock In \emph{Advances in Neural Information Processing Systems},
  volume~36.

\bibitem[{Maharana et~al.(2024)Maharana, Lee, Tulyakov, Bansal, Barbieri, and
  Fang}]{maharana2024locomo}
Adyasha Maharana, Dong-Ho Lee, Sergey Tulyakov, Mohit Bansal, Francesco
  Barbieri, and Yuwei Fang. 2024.
\newblock \href {https://doi.org/10.18653/v1/2024.acl-long.747} {{Evaluating
  Very Long-Term Conversational Memory of LLM Agents}}.
\newblock In \emph{Proceedings of the 62nd Annual Meeting of the Association
  for Computational Linguistics (Volume 1: Long Papers)}, pages 13851--13870.
  Association for Computational Linguistics.

\bibitem[{Newsome and Song(2005)}]{newsome2005taintcheck}
James Newsome and Dawn Song. 2005.
\newblock \href
  {https://www.ndss-symposium.org/ndss2005/dynamic-taint-analysis-automatic-detection-analysis-and-signaturegeneration-exploits-commodity/}
  {Dynamic taint analysis for automatic detection, analysis, and signature
  generation of exploits on commodity software}.
\newblock In \emph{Proceedings of the Network and Distributed System Security
  Symposium (NDSS)}.

\bibitem[{{OpenAI} et~al.(2023){OpenAI}, Achiam, Adler, Agarwal
  et~al.}]{achiam2023gpt4}
{OpenAI}, Josh Achiam, Steven Adler, Sandhini Agarwal, et~al. 2023.
\newblock {GPT-4} technical report.
\newblock \emph{arXiv preprint arXiv:2303.08774}.

\bibitem[{Packer et~al.(2023)Packer, Wooders, Lin, Fang, Patil, Stoica, and
  Gonzalez}]{packer2023memgpt}
Charles Packer, Sarah Wooders, Kevin Lin, Vivian Fang, Shishir~G. Patil, Ion
  Stoica, and Joseph~E. Gonzalez. 2023.
\newblock \href {https://arxiv.org/abs/2310.08560} {{MemGPT}: Towards {LLMs} as
  operating systems}.
\newblock \emph{arXiv preprint arXiv:2310.08560}.

\bibitem[{Ren et~al.(2026)Ren, Wang, Du, Xie, Liu, Yang, Feng, Pan, Zheng, Xu,
  Li, Song, Wong, and See}]{ren2026memlens}
Xiyu Ren, Zhaowei Wang, Yiming Du, Zhongwei Xie, Chi Liu, Xinlin Yang, Haoyue
  Feng, Wenjun Pan, Tianshi Zheng, Baixuan Xu, Zhengnan Li, Yangqiu Song, Ginny
  Wong, and Simon See. 2026.
\newblock \href {https://arxiv.org/abs/2605.14906} {{MemLens}: Benchmarking
  multimodal long-term memory in large vision-language models}.
\newblock \emph{arXiv preprint arXiv:2605.14906}.

\bibitem[{Shinn et~al.(2023)Shinn, Cassano, Gopinath, Narasimhan, and
  Yao}]{shinn2023reflexion}
Noah Shinn, Federico Cassano, Ashwin Gopinath, Karthik Narasimhan, and Shunyu
  Yao. 2023.
\newblock Reflexion: Language agents with verbal reinforcement learning.
\newblock In \emph{Advances in Neural Information Processing Systems
  (NeurIPS)}.

\bibitem[{Xu et~al.(2025)Xu, Liang, Mei, Gao, Tan, and Zhang}]{xu2025amem}
Wujiang Xu, Zujie Liang, Kai Mei, Hang Gao, Juntao Tan, and Yongfeng Zhang.
  2025.
\newblock \href {https://doi.org/10.52202/085713-0593} {{A-Mem}: Agentic memory
  for {LLM} agents}.
\newblock In \emph{Advances in Neural Information Processing Systems
  (NeurIPS)}, volume~38.

\bibitem[{Yao et~al.(2023)Yao, Zhao, Yu, Du, Shafran, Narasimhan, and
  Cao}]{yao2023react}
Shunyu Yao, Jeffrey Zhao, Dian Yu, Nan Du, Izhak Shafran, Karthik Narasimhan,
  and Yuan Cao. 2023.
\newblock {ReAct}: Synergizing reasoning and acting in language models.
\newblock In \emph{International Conference on Learning Representations
  (ICLR)}.

\bibitem[{Yeo et~al.(2026)Yeo, Kim, Yoon, and Hwang}]{yeo2026worldmm}
Woongyeong Yeo, Kangsan Kim, Jaehong Yoon, and Sung~Ju Hwang. 2026.
\newblock \href
  {https://openaccess.thecvf.com/content/CVPR2026/html/Yeo_WorldMM_Dynamic_Multimodal_Memory_Agent_for_Long_Video_Reasoning_CVPR_2026_paper.html}
  {{WorldMM}: Dynamic multimodal memory agent for long video reasoning}.
\newblock In \emph{Proceedings of the IEEE/CVF Conference on Computer Vision
  and Pattern Recognition (CVPR)}, pages 25599--25609.

\bibitem[{Zhang et~al.(2026)Zhang, Wang, Chen, Zhou, Wang, and
  Yan}]{zhang2026agentracer}
Guibin Zhang, Junhao Wang, Junjie Chen, Wangchunshu Zhou, Kun Wang, and
  Shuicheng Yan. 2026.
\newblock \href {https://arxiv.org/abs/2509.03312} {{AgenTracer}: Who is
  inducing failure in the {LLM} agentic systems?}
\newblock In \emph{International Conference on Learning Representations
  (ICLR)}.

\bibitem[{Zhang et~al.(2025{\natexlab{a}})Zhang, Yin, Zhang, Liu, Han, Zhang,
  Li, Wang, Wang, Chen, and Wu}]{zhang2025whowhen}
Shaokun Zhang, Ming Yin, Jieyu Zhang, Jiale Liu, Zhiguang Han, Jingyang Zhang,
  Beibin Li, Chi Wang, Huazheng Wang, Yiran Chen, and Qingyun Wu.
  2025{\natexlab{a}}.
\newblock \href {https://proceedings.mlr.press/v267/zhang25cq.html} {Which
  agent causes task failures and when? on automated failure attribution of
  {LLM} multi-agent systems}.
\newblock In \emph{Proceedings of the 42nd International Conference on Machine
  Learning}, volume 267 of \emph{Proceedings of Machine Learning Research}.

\bibitem[{Zhang et~al.(2025{\natexlab{b}})Zhang, Zhang, Xie, Li, Dai, Long,
  Xie, Zhang, Li, and Zhang}]{zhang2025gme}
Xin Zhang, Yanzhao Zhang, Wen Xie, Mingxin Li, Ziqi Dai, Dingkun Long, Pengjun
  Xie, Meishan Zhang, Wenjie Li, and Min Zhang. 2025{\natexlab{b}}.
\newblock \href
  {https://openaccess.thecvf.com/content/CVPR2025/html/Zhang_Bridging_Modalities_Improving_Universal_Multimodal_Retrieval_by_Multimodal_Large_Language_CVPR_2025_paper.html}
  {Bridging modalities: Improving universal multimodal retrieval by multimodal
  large language models}.
\newblock In \emph{Proceedings of the IEEE/CVF Conference on Computer Vision
  and Pattern Recognition (CVPR)}, pages 9274--9285.

\end{thebibliography}

\clearpage
\appendix
\renewcommand{\dbltopfraction}{0.92}
\renewcommand{\dblfloatpagefraction}{0.6}
\renewcommand{\textfraction}{0.06}
\setcounter{dbltopnumber}{3}
\makeatletter\setlength{\@fptop}{0pt}\setlength{\@dblfptop}{0pt}\makeatother
\section{Case Studies}
\label{app:cases}
We walk through failures that \method{} repairs. Each case keeps only the steps that matter for the answer, and checks and
reflections are shortened from the run logs.

\definecolor{casebar}{gray}{0.66}
\definecolor{caseback}{gray}{0.935}
\definecolor{casegood}{RGB}{30,120,60}
\definecolor{casebad}{RGB}{185,40,40}
\newcommand{\cf}[1]{\textbf{#1}\ }                       %
\newcommand{\good}[1]{\textcolor{casegood}{#1}}
\newcommand{\bad}[1]{\textcolor{casebad}{#1}}
\newcommand{\casebox}[4]{%
  \begingroup\setlength{\fboxsep}{0pt}\setlength{\fboxrule}{0.4pt}%
  \noindent\fcolorbox{casebar}{caseback}{\begin{minipage}{\dimexpr\linewidth-0.8pt\relax}%
    \colorbox{casebar}{\parbox{\linewidth}{\vspace{3pt}\hspace*{7pt}{\color{white}\small\bfseries\itshape #1}\vspace{3pt}}}\par
    \vspace{6pt}\footnotesize
    \setlength{\tabcolsep}{7pt}\renewcommand{\arraystretch}{1}%
    \begin{tabularx}{\linewidth}{@{\hspace{7pt}}>{\centering\arraybackslash}p{#2}!{\color{black!35}\vrule width 0.4pt}>{\raggedright\arraybackslash}X@{\hspace{7pt}}}
      \begin{minipage}[t]{\linewidth}\vspace{0pt}\centering #3\end{minipage} &
      \begin{minipage}[t]{\linewidth}\vspace{0pt}\raggedright #4\end{minipage}
    \end{tabularx}\par\vspace{6pt}%
  \end{minipage}}\endgroup}
\newcommand{\ph}[1]{{\rmfamily\itshape$\langle$#1$\rangle$}}
\newcommand{\promptbox}[5][System prompt.]{%
  \par\medskip\begingroup\setlength{\fboxsep}{0pt}\setlength{\fboxrule}{0.4pt}%
  \noindent\fcolorbox{casebar}{caseback}{\begin{minipage}{\dimexpr\linewidth-0.8pt\relax}%
    \colorbox{casebar}{\parbox{\linewidth}{\vspace{3pt}\hspace*{7pt}{\color{white}\small\bfseries\itshape #2}\vspace{3pt}}}\par
    \vspace{6pt}\hspace*{7pt}\begin{minipage}{\dimexpr\linewidth-14pt\relax}\raggedright\footnotesize
      \textbf{#1}\par\vspace{2pt}{\ttfamily #3\par}
      \if\relax\detokenize{#4}\relax\else
        \vspace{5pt}{\color{black!35}\rule{\linewidth}{0.4pt}}\par\vspace{3pt}
        \textbf{Input.}\par\vspace{2pt}{\ttfamily #4\par}\fi
      \if\relax\detokenize{#5}\relax\else\vspace{5pt}#5\par\fi
    \end{minipage}\par\vspace{6pt}%
  \end{minipage}}\endgroup\par\medskip}

\begin{figure*}[tp]
\casebox{Case 1 --- A missing photo label, repaired across rounds (visual search, Qwen3-VL-4B)}{0.24\linewidth}{%
\includegraphics[width=\linewidth]{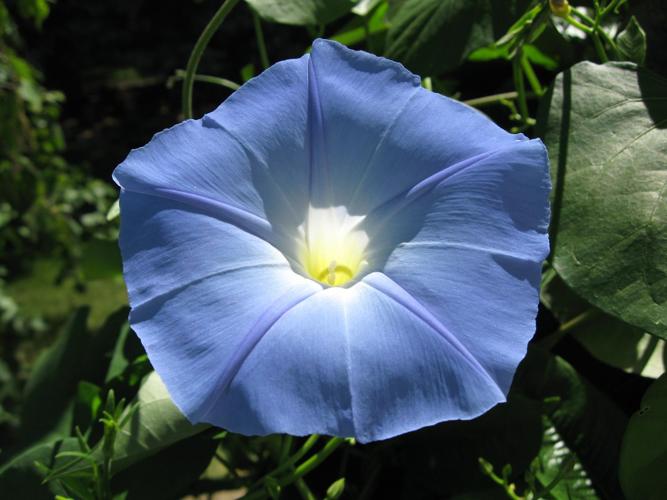}\par\vspace{1pt}
{\scriptsize Query image}\par\vspace{5pt}
\includegraphics[height=1.12cm]{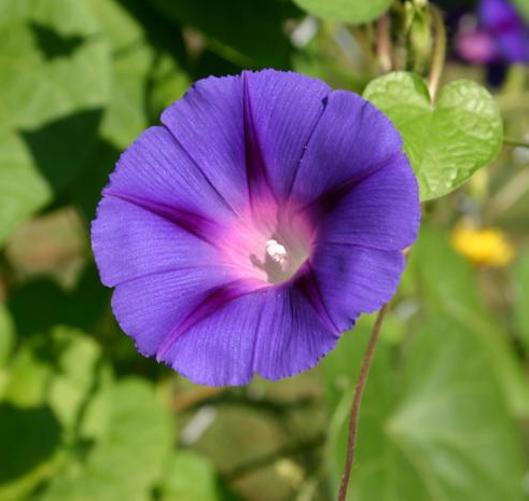}\hfill
\includegraphics[height=1.12cm]{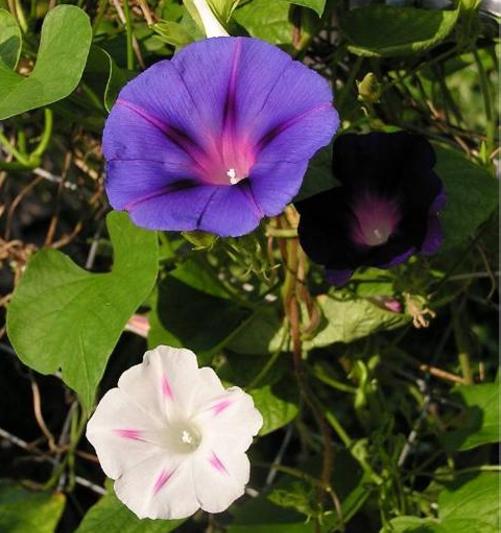}\hfill
\includegraphics[height=1.12cm]{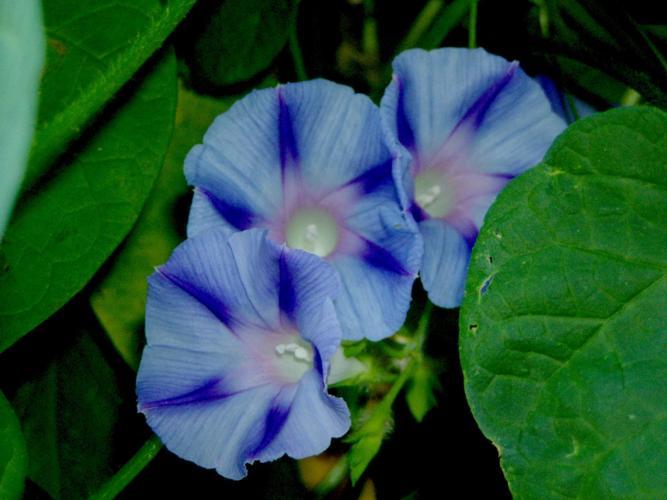}\par\vspace{1pt}
{\scriptsize D9:IMG\_001--003, kept in Memory~1}}{%
\cf{Question.} Please search for all photos of the same plant species as this image shared in the conversation.\par\vspace{2pt}
\cf{Gold Answer.} D9:IMG\_001, D9:IMG\_002, D9:IMG\_003.\par\vspace{2pt}
\cf{First Answer.} \bad{D9:IMG\_003}\par\vspace{2pt}
\cf{Memory 1 (as written).} One paragraph per turn: ``\ldots\ The photo was labeled as image D9:IMG\_001.'' /
``\ldots\ a photo of three Morning Glory flowers from the same vine \ldots'' / ``\ldots\ a photo of three Morning Glory
flowers with blue petals and dark purple stripes \ldots'' \bad{(the second and third photos are not named)}.

\smallskip
\cf{Round 1.} \emph{Check:} the one needed fact, ``the image shows a blue Morning Glory'', is confirmed by a kept image
of Memory~1; the verified facts do not answer the question. \emph{Reflection:} ``no other photos of this species were
found \ldots'', proposes \textsc{stop}. \emph{Pipeline:} \bad{rejected} (``stop needs \ldots\ every fact verified and a
failed answer from them''); rewrites Memory~1 from its three source turns: \good{``Aiden shared three photos of Morning
Glory flowers \ldots\ labeled D9:IMG\_001, D9:IMG\_002, and D9:IMG\_003.''} \emph{Answer:} \bad{D9:IMG\_003}.

\smallskip
\cf{Round 2.} \emph{Check} (with the note of round~1): ``the plant species in the query image'' $\rightarrow$
``Morning Glory''; ``all photos of the same plant species'' $\rightarrow$ the rewritten sentence of Memory~1.
\emph{Reflection:} the facts ``clearly indicate that the answer should include all three photos'', proposes
\stage{Answer} from Memory~1. \emph{Pipeline:} \good{accepted}. \emph{Answer:} \good{D9:IMG\_001, D9:IMG\_002, D9:IMG\_003}.

\smallskip
\cf{Round 3.} The same facts are verified in Memory~1; a new answer listing only D9:IMG\_003 is not taken, since it says less.}
\caption{Case 1 (Mem-Gallery, Nature\_Economics\_Programming\_Student\_Life\#106). Checks and reflections are quoted
from the run log. }
\label{fig:case1}
\end{figure*}

\paragraph{Case 1: a missing photo label, repaired across rounds.}
Memory~1 holds the three turns in which Aiden shares photos of Morning Glory flowers, but only the text of the first
turn names its photo (Figure~\ref{fig:case1}), so the system answers with one photo. In round~1 the model checks only the
species in the query image. Its reflection notes that no other photos of this species were found and proposes to stop,
which the pipeline rejects because no answer has yet been tried from verified facts; the pipeline instead rewrites
Memory~1 from its source turns, and the new text names all three photos. With the note of round~1, round~2 looks for
all photos of the species and finds them in the rewritten Memory~1, and round~3 verifies the same facts and keeps the
answer.

\begin{figure*}[tp]
\casebox{Case 2 --- A misreading, fixed without touching memory (conflict detection, Qwen3-VL-4B)}{0.24\linewidth}{%
\includegraphics[width=0.82\linewidth]{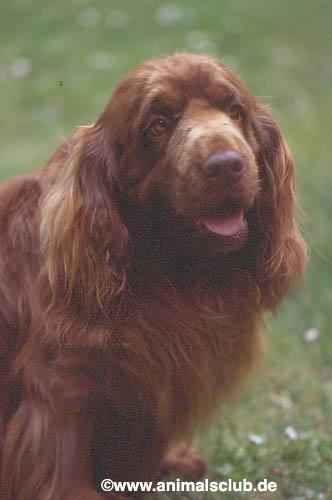}\par\vspace{1pt}
{\scriptsize D2:IMG\_001: Rocky, a Sussex Spaniel}\par\vspace{5pt}
\includegraphics[width=0.82\linewidth]{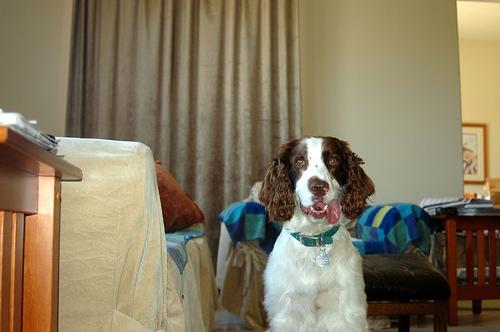}\par\vspace{1pt}
{\scriptsize D2:IMG\_005: Luna}}{%
\cf{Question.} Please determine whether the following information conflicts with the dialogue memory: Sussex Spaniel has
white-and-liver coat pattern and alert eyes.\par\vspace{2pt}
\cf{Gold Answer.} Yes.\par\vspace{2pt}
\cf{First Answer.} \bad{No.}\par\vspace{2pt}
\cf{Memory 1.} ``The assistant suggested identifying Luna by her white-and-liver coat pattern and alert eyes, noting that
Sussex Spaniels are solid golden-brown and heavier in build.''

\smallskip
\cf{Round 1.} \emph{Check:} the statement is located in Memory~1, in the sentence above. \emph{Reflection:} the verified fact ``contradicts the
statement by asserting that Sussex Spaniels are solid golden-brown \ldots\ the answerer misread or ignored the
conflicting fact'', proposes \stage{Answer} from Memory~1. \emph{Pipeline:} \good{accepted}; answers from the verified
facts only. \emph{Answer:} \good{Yes.}

\smallskip
\cf{Round 2.} The check verifies ``Sussex Spaniels are solid golden-brown'' in Memory~1; nothing is found wrong, and the
answer is kept.}
\caption{Case 2 (Mem-Gallery, Dog\_Behavior\_Research\_Academic\_Life\#10).}
\label{fig:case2}
\end{figure*}

\paragraph{Case 2: a misreading, fixed without touching memory.}
The statement in the question gives Luna's coat to the Sussex Spaniel, and the system misses the conflict
(Figure~\ref{fig:case2}). Memory~1 is correct: the check finds the deciding sentence there. The reflection recognizes
that the answer misread this sentence rather than that memory is wrong, and proposes to answer again from the verified
facts; the pipeline accepts, and the answer is right in round~1 while Memory~1 stays unchanged.

\paragraph{Case 3: an injected omission, located at \stage{Write}.}
This failure comes from the propagation analysis of Sec.~\ref{sec:propagation}, where we remove a value from a record
and therefore know the source stage (Figure~\ref{fig:case3}). After the injection, Memory~1 says that warehouse
positioning cuts lead time ``by a significant amount'', and the system answers exactly that. In the first round, which
receives no feedback, the model's own reflection finds nothing wrong, since the answer matches the record. The pipeline
does not accept this: a separate check finds that the record does not tell \emph{how much}, so it traces the error to
\stage{Write}, rewrites Memory~1 from the original conversation, and the value returns.

\begin{figure*}[tp]
\casebox{Case 3 --- An injected omission, located at Write (factual retrieval, Qwen3-VL-30B-A3B)}{0.24\linewidth}{%
\raggedright
\cf{Original turn.} \emph{User:} ``How important is warehouse positioning in cross-border e-commerce?''
\emph{Assistant:} ``Critical. Proximity to airports and bonded zones can cut lead time by \good{20--40\%}.''\par\vspace{6pt}
\cf{Memory 1 after the injection.} ``\ldots\ The assistant responded that it is critical, noting that proximity to
airports and bonded zones can cut lead time by \bad{a significant amount}.''}{%
\cf{Question.} How much can warehouse positioning near airports and bonded zones reduce lead time in cross-border
e-commerce?\par\vspace{2pt}
\cf{Gold Answer.} 20--40\%.\par\vspace{2pt}
\cf{Injected error.} The value is removed from Memory~1 when it is written.\par\vspace{2pt}
\cf{First Answer.} \bad{A significant amount.}

\smallskip
\cf{Round 1} (no feedback). \emph{Check:} Memory~1 holds the needed statement, ``\ldots\ can cut lead time by a
significant amount'', but it does not say how much, and no other record says more. \emph{Reflection:} ``The answer
`A significant amount' is supported by the fact in Memory~1 \ldots\ No further evidence is needed''; it proposes no
repair. \emph{Pipeline:} \bad{rejected}, since the record does not answer the question; it traces the error to
\stage{Write} and rewrites Memory~1 from the original turn: \good{``\ldots\ can cut lead time by 20--40\%.''}
\emph{Answer:} \good{20--40\%}.}
\caption{Case 3 (Mem-Gallery, Entrepreneurship\_Blockchain\_Economics\_Logistics\_Nature\#38; an omission injected as
in Sec.~\ref{sec:propagation}). The error is located and repaired in the first round, before any feedback.}
\label{fig:case3}
\end{figure*}

\section{Experimental Details and Additional Analyses}
\label{app:details}

\subsection{Why Stage-wise Diagnosis}
\label{app:oneshot}
Our first diagnoser followed trajectory-level reflection~\citep{shinn2023reflexion}: told only that its answer was
wrong, the model read the whole memory trace of the question in one call, named the faulty stage and record (prompt
below), and the repair of that stage was applied. On 215 failures (the 30B model on
the original ten Mem-Gallery scenes), this diagnosis collapsed onto the answer stage: 74\% of its first-round diagnoses
named \stage{Answer}, and its repairs recovered no more failures than answering again with step-by-step analysis
(32\% vs.\ 31\% of the failures that are not yes/no questions). A first stage-wise version, which checked one stage at
a time, recovered 60\%, so we built \method{} on stage-wise checks (Sec.~\ref{sec:memdoctor}).

\begin{figure*}[tp]
\centering
\includegraphics[width=\textwidth]{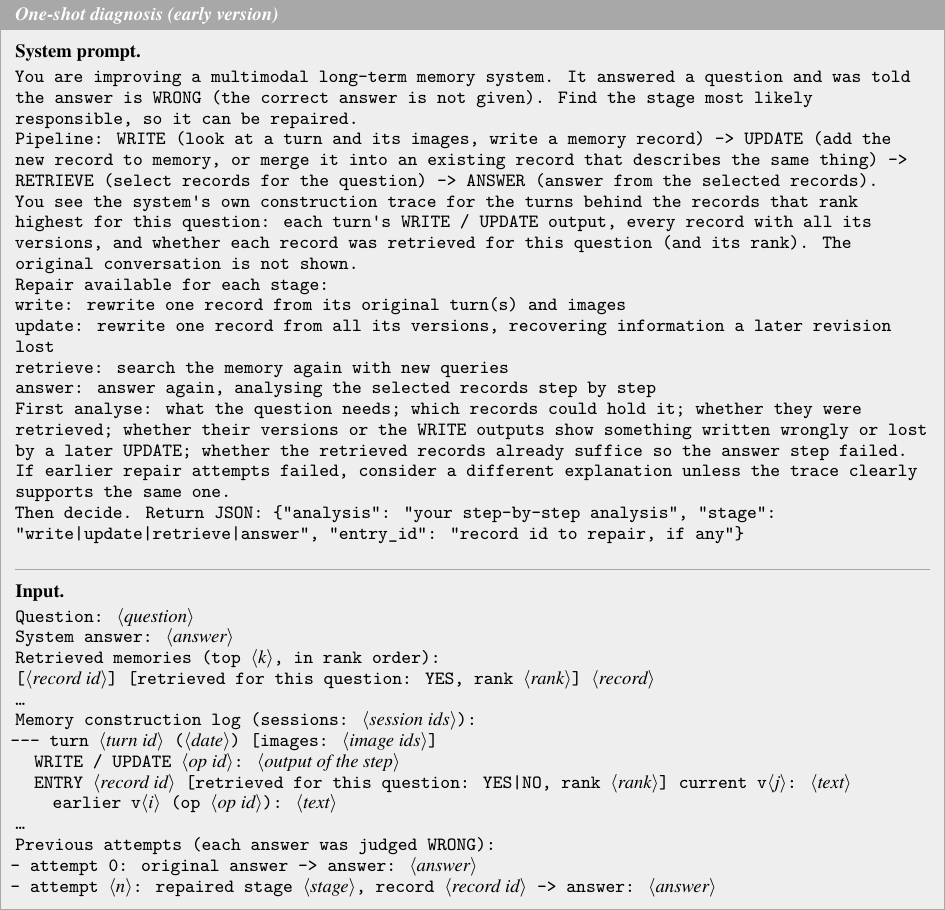}
\end{figure*}

\subsection{Feedback Protocol for Locating Injected Errors}
\label{app:feedback}
Section~\ref{sec:propagation_results} runs self-diagnosis with outcome feedback on 247 wrong answers whose source stage
is known. A question enters this set if the clean system answers it correctly, it is asked on an edited turn, it is
answered wrongly after the injection but not after the control edit, and the injected error is still in memory at the
end of the dialogue. Retrieval misses change no memory and have no control, so they enter whenever they are answered
wrongly. Most questions are the targets of their injections; the rest (6\%) are other questions on the edited turns.

\begin{figure*}[tp]
\centering
\includegraphics[width=\linewidth]{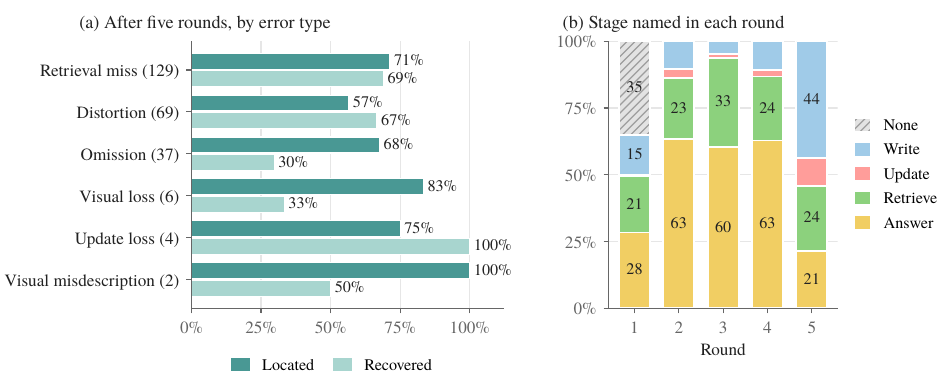}
\caption{Self-diagnosis with outcome feedback on the 247 injected failures (30B system, at most five rounds).
(a)~Share of the failures of each error type whose source stage is named in some round (\emph{located}) and that are
answered correctly again (\emph{recovered}); the number of failures is in parentheses. (b)~Stage named in each round,
as a share of the questions that reach that round (247, 180, 144, 121 and 98); \emph{none} (nothing found wrong) is
possible only in the first round, which has no feedback.}
\label{fig:feedback_rounds}
\end{figure*}

The first round has no feedback; from the second on, the model is told only that the earlier answers were judged
wrong, and a run stops once an answer is judged correct or after five rounds. Figure~\ref{fig:feedback_rounds} shows
the results. Of the three frequent error types, retrieval misses are located and recovered most often (71\% and
69\%). Omissions are often located but
rarely recovered (68\% vs.\ 30\%), whereas distortions are recovered more often than they are located (67\% vs.\
57\%). Panel~(b) shows the diagnosis moving upstream: without feedback, a third of the failures look correct to the
model; once told that the answer is wrong, it first suspects the answering step (about 60\% of the diagnoses in rounds
2--4), and only in the fifth round does \stage{Write} become the most frequent diagnosis (44\%).

\subsection{Propagation Details}
\label{app:propagation}
This section gives the details of the propagation analysis (Sec.~\ref{sec:propagation}): its results split by
redundancy, how an edit is accepted, the control edits, the retrieval errors and the editing prompts. Table~\ref{tab:redundancy} splits the
accepted injections of Table~\ref{tab:propagation} by whether the memory can still answer the target question after the
edit (\emph{redundant}, defined below). Each injection is paired with its control on the same target question:
\emph{wrong answer} is the share of target questions answered wrong after the injection and after the control, and
$\Delta$ is the difference between the two in points. An error rarely changes the answer when other records or the
attached images still hold the fact, and often does when the edited records were its only evidence.

\begin{table*}[tp]
\centering
\small
\begin{tabular}{lcrcc}
\toprule
Error & Redundant & $N$ & Wrong answer (inj.\ / ctrl.) & $\Delta$ \\
\midrule
Omission              & no  & 49 & 61\% / 2\%  & +59  \\
                      & yes & 81 & 9\% / 4\%   & +5   \\
Distortion            & no  & 75 & 71\% / 3\%  & +68  \\
                      & yes & 58 & 12\% / 2\%  & +10  \\
Visual loss           & no  & 10 & 40\% / 0\%  & +40  \\
                      & yes & 2  & 100\% / 0\% & +100 \\
Visual misdescription & no  & 4  & 75\% / 0\%  & +75  \\
                      & yes & 8  & 12\% / 0\%  & +12  \\
Update loss           & no  & 5  & 80\% / 0\%  & +80  \\
                      & yes & 6  & 17\% / 0\%  & +17  \\
\bottomrule
\end{tabular}
\caption{The injections of Table~\ref{tab:propagation} split by redundancy. \emph{Redundant}: after the edit, other
records or the images still attached still answer the target question. $\Delta$: paired difference in points.}
\label{tab:redundancy}
\end{table*}

\paragraph{Acceptance checks.}
An edit is made to the records written at the evidence turns, before any later merge, and is kept only if it passes
every check below. Each probe asks the system's own answering step and scores the answer with the judge.
(i)~Before the edit, the clean evidence records alone answer the target question.
(ii)~The edit changes something and still reads as an ordinary record: it adds no more than three new words, or 12\%
of its words if that is more, adds no negation, does not
say that something was removed or is unknown, and does not paste the dialogue as ``speaker: \ldots''. An omission keeps
at least half of the record's words and leaves no placeholder such as ``something'' or ``a certain''. An edited image
label gains at most three new words.
(iii)~The edited evidence records alone no longer answer the question, and a distortion makes them give the planted
wrong value. For an omission and a visual misdescription, this probe reads the text only, so that an attached image
cannot hide the error.
(iv)~Every other question on the same turns that the clean records answer is still answered from the edited records; a
question that asks for the same fact counts as a second target.
An update loss is checked the same way on the merged record, read together with the other evidence records: the version
before the merge and the clean merged version both answer the question, the edited merge does not, it keeps at least
80\% of the words that came from the new information, and the merge guard of our system would not catch the loss. If an
omission or an update loss still fails after the retry, every sentence that contains the answer is deleted instead and
the checks are run again.

\paragraph{Control edits.}
All injections on the same evidence turns share one control. The control reverses the order of the sentences in each
evidence record and changes nothing else; a sentence that begins with a pronoun or a connective (``She'', ``Then'')
stays with the one before it, and an abbreviation does not end a sentence, so that references and time order are kept.
A record with a single sentence is prefixed with ``Noted:'' instead. The control must still answer every target
question paired with it; otherwise it is replaced by the ``Noted:'' prefix (2\% of the controls). For an update loss, the
control reorders the merged record of the same merge. The control is replayed exactly like the injection, through every
later merge, and each injection is compared with its own control on the same target question.

\paragraph{Redundant injections.}
After an edit is accepted, the system answers the target question from the clean top-$k$ records, with the records of
the evidence turns replaced by their edited versions. If this answer is still right, other records or the images still attached give the fact, and
the injection is flagged as \emph{redundant}. Redundant injections stay in Table~\ref{tab:propagation}; they are 62\% of
the omissions and 44\% of the distortions. Table~\ref{tab:redundancy} splits each error type by this flag. Without
redundancy, an omission raises the rate of wrong answers over its control by 59 points and a distortion by 68; with
redundancy, by 5 and 10. The other types have 2 to 10 injections per group.

\paragraph{Retrieval errors.}
The two retrieval errors change no memory and are not replayed: the target question is answered once more under the
changed ranking. Their target questions are those the clean system answers correctly, other than yes/no questions,
refusals and questions with an image, at most 16 per scene. For a retrieval miss, the evidence records are pushed
below the top-$k$, to a random lower rank. For a retrieval distraction, the
evidence stays in the top-$k$, and the two non-evidence records below the top-$k$ that are most similar to the question
are lifted to the top two ranks; records that contain the answer are never lifted. Its control lifts the two least
similar records in the same way, and a question is kept only if both apply.

\paragraph{Editing prompts.}
For the record edits in Sec.~\ref{sec:propagation}, we use Qwen3-VL-30B-A3B-Instruct with the prompts listed in
Appendix~\ref{app:propagation_prompts}. An edit that fails a check is sent back once with the message
{\ttfamily Your edit was rejected: \ph{reasons}. Edit the ORIGINAL records again and fix this.}

\subsection{Baselines and MemLens Setup}
We evaluate six baselines---TextRAG, MuRAG, A-Mem, Mem0, Omni-SimpleMem, and V-Mem---on the 195-question agent subset of MemLens at the 32K history length.
All six use Qwen3-VL-8B-Instruct for answering and any LLM-based memory construction.
Each question is processed independently from its associated history, preserving session order and timestamps.
TextRAG indexes raw dialogue turns with BGE-M3, while our MuRAG adaptation uses GME-Qwen2-VL-2B-Instruct to rank turns by fused text--image representations.
For A-Mem, Mem0, and Omni-SimpleMem, the same 8B model generates image descriptions from the images and their local dialogue context before memory construction.
V-Mem retains its multimodal indexes and modality-aware retrieval.
The baselines share a final-answer prompt and use a retrieval budget of ten turns or memory records, except for V-Mem, which uses its native retrieval budget of twelve; at most four retrieved images are passed to the answerer where available.
Our system uses Qwen3-VL-4B-Instruct and retains its Mem-Gallery settings, including six retrieved records.
All answers are scored by Qwen3-VL-30B-A3B-Instruct using MemLens's official task-specific rubrics.

\section{\method{} on Existing Memory Backends}
\label{app:backend_repair}

\subsection{Choice of Backends}
We attach \method{} to MuRAG and A-Mem because together they span the stages of
our framework (Sec.~\ref{sec:problem_formulation}). Our adapted
MuRAG~\citep{chen2022murag} stores raw dialogue turns and supports the
\stage{Retrieve} and \stage{Answer} stages. It ranks turns by fused text--image
representations and answers from them. We use this backend to evaluate
stage-guided re-retrieval and re-answering. A-Mem~\citep{xu2025amem} has all four
stages: it writes structured notes, evolves existing notes and their links as
new information arrives (its counterpart of \stage{Update}, though not a record
merge), and retrieves notes for answering, so it also tests local repair of the
memory itself. Its visual inputs use the caption-to-text adaptation described
above.

The adapters expose only the operations each backend supports. MuRAG permits
retrieval and answer repair; A-Mem additionally permits source-backed writing
and update repair when the required provenance and version history are
available. For A-Mem, an accepted repair reconstructs the target note from its
authorized sources, runs native note analysis and re-embedding, and preserves
its effective links without triggering a new cascade of memory evolution.
Changes are local to the current question. Thus, the comparison tests the
applicability of \method{} across different backend capabilities,
rather than assuming that both backends implement an identical pipeline.

\subsection{Evaluation Protocol}
We use the same 195-question MemLens subset at the 32K history length
as in the main evaluation~\citep{ren2026memlens}. For this backend-attachment
experiment, all diagnostic, reflection, and repair inference uses the original
Qwen3-VL-8B-Instruct model. We reuse each backend's initial memories and paired baseline 
answers and retain its native retrieval, context construction, and answering implementation. A-Mem's paired
baseline is denoted $B^\prime$: its answers are tied to the instrumented,
verified initial note state and version history used by both Reflexion and \method{}.
It is distinct from the earlier A-Mem baseline in the main system comparison.

We compare the original answers, Reflexion~\citep{shinn2023reflexion}, and \method{}.
Reflexion and \method{} receive the same trace representation and share the
non-stage control rules. Reflexion rewrites queries and
re-answers without stage localization or memory editing. \method{} selects
among the backend's supported repair actions. Every question, including those
with an initially correct answer, can undergo up to three rounds. Gold answers
and correctness labels are unavailable during these rounds. After inference,
the original 30B judge scores the saved trajectories using the native MemLens
task-specific protocol.

To integrate \method{} with MuRAG and A-Mem, we adapt the diagnosis inputs
and repair interface to each backend's memory representation and supported
operations.

\subsection{Results on MemLens}
\providecommand{\brup}[1]{{\fontsize{5pt}{6pt}\selectfont\textcolor[RGB]{0,128,60}{+#1}}}
\providecommand{\brdown}[1]{{\fontsize{5pt}{6pt}\selectfont\textcolor[RGB]{190,30,30}{$-$#1}}}
\providecommand{\brzero}{{\fontsize{5pt}{6pt}\selectfont\textcolor{gray}{0.00}}}
\providecommand{\brd}[1]{\hspace{0.5pt}\raisebox{-0.5ex}{#1}}
\begin{table*}[t]
\centering
\small
\setlength{\tabcolsep}{0pt}
\begin{tabular}{l@{\hspace{14pt}}*{5}{r@{}l@{\hspace{14pt}}}r@{}l}
\toprule
Procedure & \multicolumn{2}{c}{IE (61)} & \multicolumn{2}{c}{KU (29)} & \multicolumn{2}{c}{TR (48)} & \multicolumn{2}{c}{MSR (35)} & \multicolumn{2}{c}{AR (22)} & \multicolumn{2}{c}{Overall (195)} \\
\midrule
\multicolumn{13}{@{}l}{MuRAG} \\
Original $B$ & 39.34 &  & 31.03 &  & \underline{52.08} &  & 17.14 &  & 100.00 &  & 44.10 &  \\
Reflexion & \underline{50.82} & \brd{\brup{11.48}} & \textbf{37.93} & \brd{\brup{6.90}} & \underline{52.08} & \brd{\brzero} & 17.14 & \brd{\brzero} & 100.00 & \brd{\brzero} & \underline{48.72} & \brd{\brup{4.62}} \\
\method{} & \textbf{59.02} & \brd{\brup{19.67}} & \textbf{37.93} & \brd{\brup{6.90}} & \textbf{54.17} & \brd{\brup{2.08}} & 17.14 & \brd{\brzero} & 100.00 & \brd{\brzero} & \textbf{51.79} & \brd{\brup{7.69}} \\
\midrule
\multicolumn{13}{@{}l}{A-Mem} \\
Original $B^\prime$ & 44.26 &  & 31.03 &  & \underline{45.83} &  & 14.29 &  & \textbf{100.00} &  & 43.59 &  \\
Reflexion & \underline{47.54} & \brd{\brup{3.28}} & \underline{34.48} & \brd{\brup{3.45}} & \textbf{47.92} & \brd{\brup{2.08}} & 14.29 & \brd{\brzero} & \textbf{100.00} & \brd{\brzero} & \underline{45.64} & \brd{\brup{2.05}} \\
\method{} & \textbf{52.46} & \brd{\brup{8.20}} & \textbf{48.28} & \brd{\brup{17.24}} & 43.75 & \brd{\brdown{2.08}} & 14.29 & \brd{\brzero} & 95.45 & \brd{\brdown{4.55}} & \textbf{47.69} & \brd{\brup{4.10}} \\
\bottomrule
\end{tabular}
\caption{Accuracy (\%) on the same MemLens195 questions after attaching Reflexion or \method{}
to 8B backends. IE: information extraction; KU: knowledge update; TR:
temporal reasoning; MSR: multi-session reasoning; AR: answer refusal.
The numbers in parentheses are question counts. $B^\prime$ is the paired
instrumented A-Mem baseline. Small numbers give the change from the original
answers ($B$ or $B^\prime$), in points. \textbf{Bold} and \underline{underline}
mark the highest and second-highest accuracy within each backend; ties share the
mark, and a column where all three tie is left unmarked.
Each value comes from one observed run of the original attachment protocol.}
\label{tab:backend_repair_categories}
\end{table*}

\begin{table*}[t]
\centering
\small
\setlength{\tabcolsep}{2pt}
\begin{tabular*}{\linewidth}{@{\extracolsep{\fill}}l*{10}{r@{}l}@{}}
\toprule
Procedure & \multicolumn{2}{c}{FR} & \multicolumn{2}{c}{MR} & \multicolumn{2}{c}{TR} & \multicolumn{2}{c}{VR} & \multicolumn{2}{c}{TTL} & \multicolumn{2}{c}{VS} & \multicolumn{2}{c}{CD} & \multicolumn{2}{c}{KR} & \multicolumn{2}{c}{AR} & \multicolumn{2}{c}{Overall} \\
\midrule
\multicolumn{21}{@{}l}{MuRAG} \\
Original $B$ & 71.69 &  & 78.16 &  & 69.92 &  & 49.43 &  & \textbf{85.76} &  & 75.82 &  & 62.96 &  & \underline{62.96} &  & \textbf{94.02} &  & 75.16 &  \\
Reflexion & \underline{72.15} &  & \textbf{80.10} &  & \textbf{74.80} &  & \textbf{57.47} &  & 85.16 &  & \underline{77.12} &  & \underline{65.43} &  & \underline{62.96} &  & 86.96 &  & \underline{76.10} &  \\
\method{} & \textbf{73.52} & \brd{\brup{1.83}} & \underline{78.64} & \brd{\brup{0.49}} & \underline{73.17} & \brd{\brup{3.25}} & \underline{51.72} & \brd{\brup{2.30}} & \underline{85.46} & \brd{\brdown{0.30}} & \textbf{78.10} & \brd{\brup{2.29}} & \textbf{66.67} & \brd{\brup{3.70}} & \textbf{66.67} & \brd{\brup{3.70}} & \textbf{94.02} & \brd{\brzero} & \textbf{76.62} & \brd{\brup{1.46}} \\
\midrule
\multicolumn{21}{@{}l}{A-Mem} \\
Original $B^\prime$ & \underline{69.41} &  & 60.68 &  & 68.29 &  & 51.72 &  & \underline{63.20} &  & 66.01 &  & 32.10 &  & \underline{71.60} &  & \textbf{95.11} &  & 65.75 &  \\
Reflexion & 68.49 &  & \textbf{64.08} &  & \textbf{78.86} &  & \textbf{58.62} &  & \underline{63.20} &  & \textbf{68.63} &  & \textbf{34.57} &  & 70.37 &  & 92.39 &  & \textbf{67.74} &  \\
\method{} & \textbf{69.86} & \brd{\brup{0.46}} & \underline{62.14} & \brd{\brup{1.46}} & \underline{73.17} & \brd{\brup{4.88}} & \underline{54.60} & \brd{\brup{2.87}} & \textbf{64.09} & \brd{\brup{0.89}} & \underline{67.97} & \brd{\brup{1.96}} & \underline{33.33} & \brd{\brup{1.23}} & \textbf{74.07} & \brd{\brup{2.47}} & \underline{94.57} & \brd{\brdown{0.54}} & \underline{67.27} & \brd{\brup{1.52}} \\
\bottomrule
\end{tabular*}
\caption{Accuracy (\%) on all 1,711 Mem-Gallery questions after attaching
Reflexion or \method{} to 8B backends. $B$ and $B^\prime$ denote the paired MuRAG
and A-Mem baselines, respectively. Overall accuracy is question-weighted.
\textbf{Bold} and \underline{underline} mark first and second place within
each backend and column. Tied scores share a rank. The small numbers next to
\method{} give the change from the paired baseline in percentage points.}
\label{tab:backend_repair_mg}
\end{table*}

Table~\ref{tab:backend_repair_categories} reports all five question types.
\method{} improves overall MuRAG accuracy from 44.10\% to 51.79\%
($+7.69$ points) and A-Mem from 43.59\% to 47.69\% ($+4.10$ points).
The corresponding Reflexion results are 48.72\% and 45.64\%.
The gains concentrate in information extraction and knowledge update; the other
three types change little.

\paragraph{Information extraction (IE).}
Every IE question in this subset has an image among its evidence, and most ask about a detail in it.
In 32 of MuRAG's 37 initial failures and 30 of A-Mem's 34, the first diagnosis of
\method{} finds the needed fact in the image behind a retrieved record: the
evidence was retrieved, but the backend answered ``Not mentioned'' or misread it. Checking each needed fact against the record and its
image recovers 12 of these MuRAG failures and 6 of the A-Mem ones, on A-Mem sometimes after rewriting the
note from its source. Reflexion reconsiders the answer as a whole without
pointing to the image that holds the fact, and recovers 7 and 2.

\paragraph{Knowledge update (KU).}
KU asks for the current value of something the user has changed. On A-Mem, the
backend answers 9 of its 20 failures with the old value (``fedora'' for a
favorite hat that is now a baseball cap) and 11 with ``Not mentioned''; in 13 of
them the first diagnosis finds no retrieved note with the current value.
\method{} then searches for it with a query tied to the question date (``favorite
bottle user 2024-06-01''), which recovers four, and answering again recovers a
fifth. Reflexion recovers one of these five and repeats the same answer in every
round on the other four. 

\paragraph{Other question types (TR, MSR, AR).}
On temporal reasoning, both methods stay within one question of the original
accuracy on both backends; half of these questions (25 of 48) ask which of two
periods lasted longer, which needs the start and end of both periods rather than
a single located fact. On multi-session reasoning, accuracy is unchanged under
both methods, since a count or a sum over several sessions is stated in no single
record. All 22 answer-refusal questions are answered correctly before repair;
both methods keep all of them on MuRAG, and on A-Mem Reflexion keeps 22 and
\method{} 21.

\paragraph{Summary.}
Localization and repair carry over to both backends. \method{} raises MuRAG to
51.79\% and A-Mem to 47.69\%, above Reflexion on both (48.72\% and 45.64\%), and
recovers 19 and 13 initial failures against 10 and 5. On Mem-Gallery
(App.~\ref{app:backend_repair_mg}), \method{} also improves both backends (+1.46
points on MuRAG and +1.52 on A-Mem), ahead of Reflexion on MuRAG and within 0.5
points of it on A-Mem. On MuRAG both methods use the same retrieval and
answering operations. The evaluated Loop also conditions its answerer on
reflection analysis; the comparison therefore measures the combined effects of
action selection and answer conditioning, rather than isolating stage
localization. On A-Mem the repairs follow the located
stage: rewriting a note from its source (\stage{Write}), searching for the newer
note a KU question needs (\stage{Retrieve}), and re-answering from the selected
records (\stage{Answer}). \stage{Update} repairs are issued as well, but A-Mem stores
a changed value as a new note and its evolution revises only the context, tags
and links of older notes, so an outdated answer comes from a retrieval that
misses the newer note.

\subsection{Results on Mem-Gallery}
\label{app:backend_repair_mg}
We further evaluate attachment to MuRAG and A-Mem on all 1,711 Mem-Gallery
questions. Both repair methods start from the same paired baseline within each
backend. The original 30B judge scores the saved answers using the native
Mem-Gallery evaluation protocol. Table~\ref{tab:backend_repair_mg} reports
final-answer accuracy across all nine question types.

The paired A-Mem baseline in this repair evaluation is 65.75\%, compared with
66.80\% in the main comparison. Both repair methods start from the same initial
memories and baseline answers, and all improvements are measured relative to
this paired baseline.

\method{} improves MuRAG from 75.16\% to 76.62\% ($+1.46$ points), compared
with 76.10\% for Reflexion. On A-Mem, it improves the paired baseline from
65.75\% to 67.27\% ($+1.52$ points), while Reflexion reaches 67.74\%.

The comparison also varies by question type. \method{} achieves higher accuracy
than Reflexion on factual retrieval and knowledge resolution with both backends.
It also preserves MuRAG's answer-refusal accuracy at 94.02\%, retaining all 173
initially correct refusal answers, while Reflexion breaks 13 of them. On A-Mem,
\method{} breaks one initially correct refusal answer, compared with five under
Reflexion. By contrast, Reflexion achieves higher accuracy on temporal and
visual-centric reasoning with both backends.

\onecolumn %
\section{Prompts of \method{}}
\label{app:prompts}

The prompts are ordered as one round of \method{} (Sec.~\ref{sec:memdoctor}) uses them: the checks that locate the error
from the answer backward (App.~\ref{app:prompts_locate}), the reflection that proposes the next step
(App.~\ref{app:prompts_reflect}), and the repairs followed by a new answer (App.~\ref{app:prompts_repair}). All are listed
verbatim; only the two location labels of Prompt~1, \texttt{text} and \texttt{missing}, are renamed for readability.
Each is the system message of one model call, and the inputs below it form the user message; every record carries its
id, and images follow the record that keeps them.

\subsection{Locating the Error, from the Answer Backward}
\label{app:prompts_locate}
\stage{Write} needs no prompt of its own: a needed fact that none of these checks finds stated correctly points to it.
\par\nopagebreak %

\par\medskip\noindent\includegraphics[width=\linewidth]{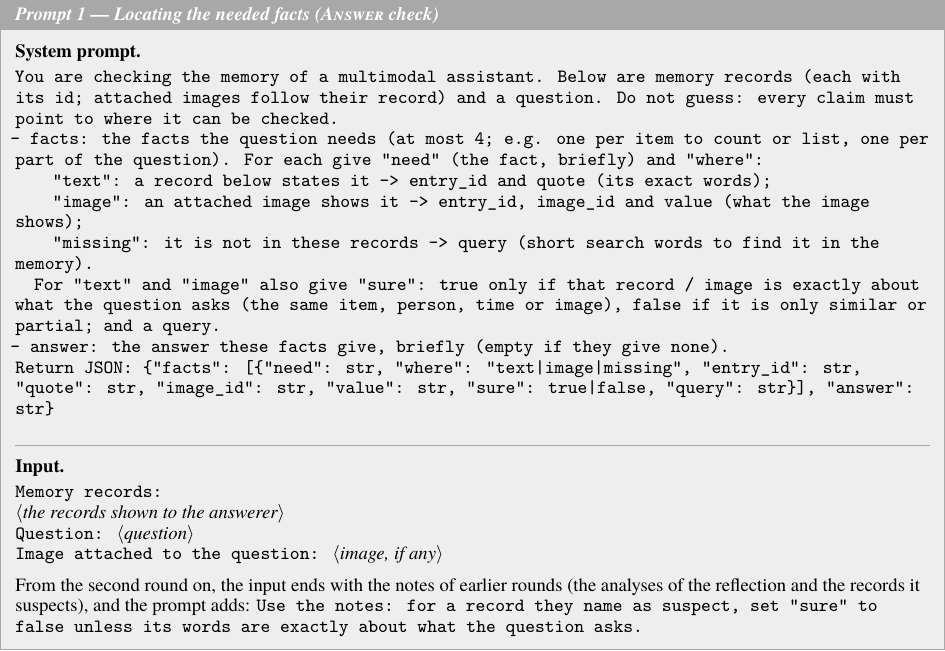}\par\medskip

\par\medskip\noindent\includegraphics[width=\linewidth]{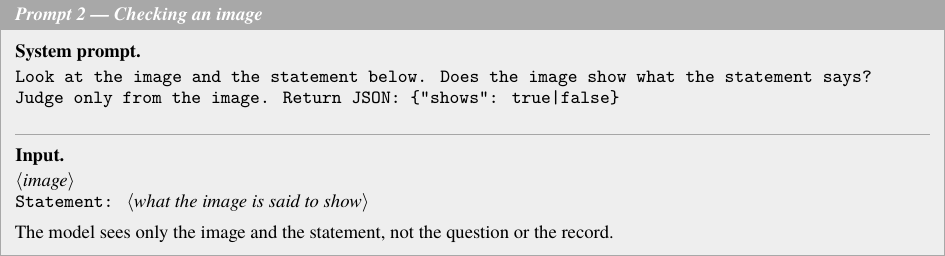}\par\medskip

\par\medskip\noindent\includegraphics[width=\linewidth]{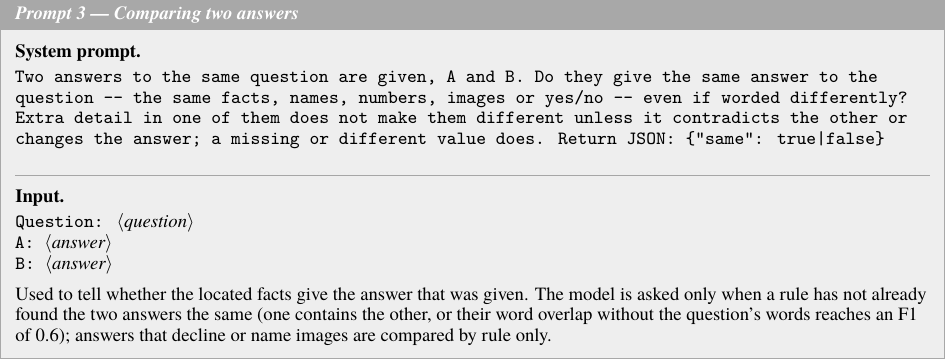}\par\medskip

\par\medskip\noindent\includegraphics[width=\linewidth]{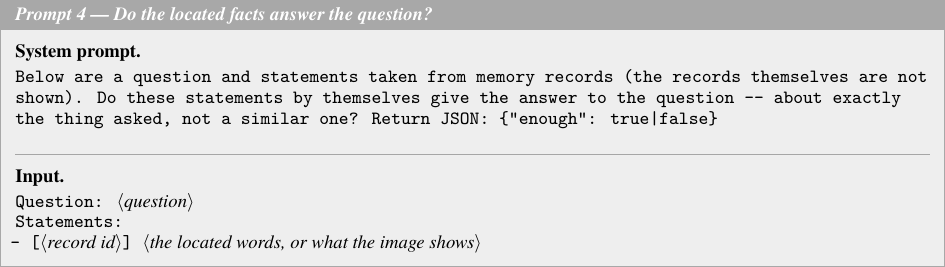}\par\medskip

\par\medskip\noindent\includegraphics[width=\linewidth]{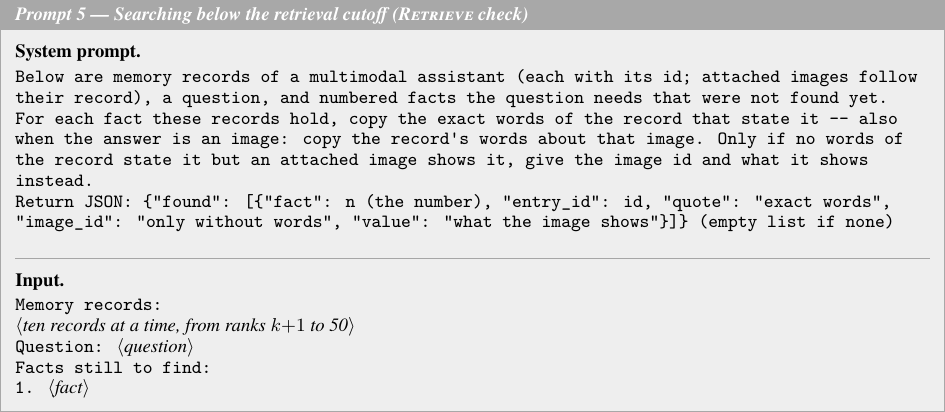}\par\medskip

\par\medskip\noindent\includegraphics[width=\linewidth]{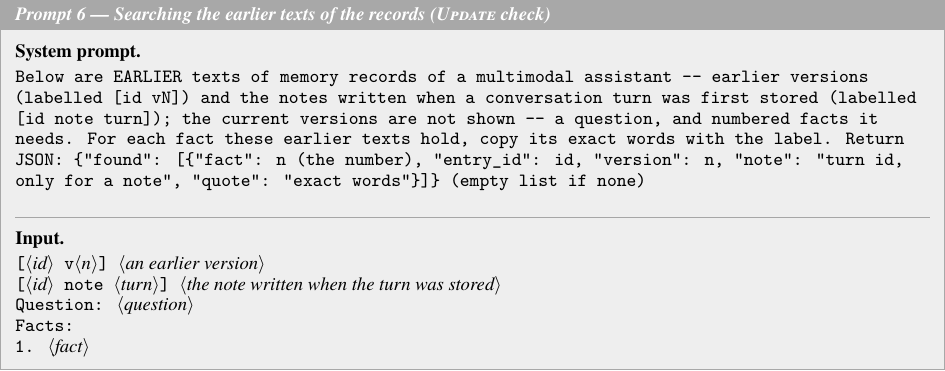}\par\medskip

\subsection{Reflection Across Rounds}
\label{app:prompts_reflect}

\par\medskip\noindent\includegraphics[width=\linewidth]{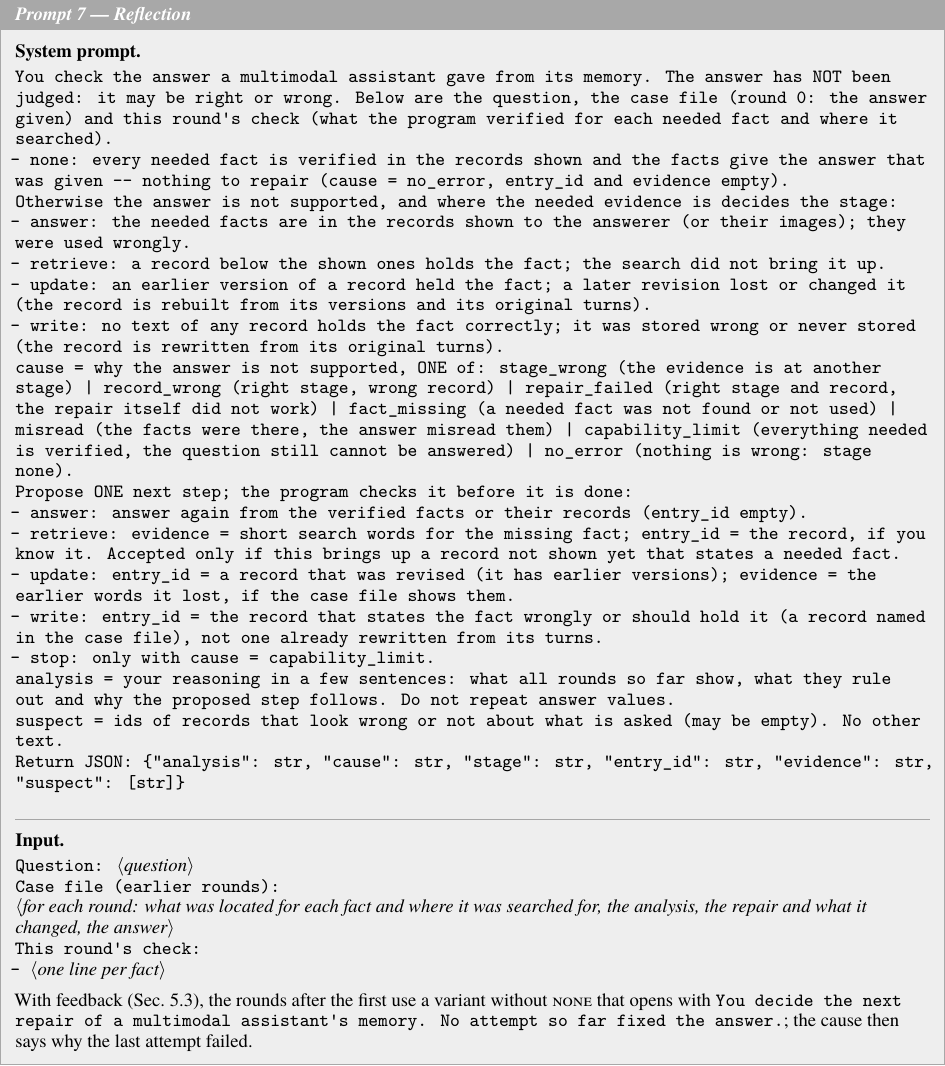}\par\medskip

\subsection{Repair and Answering Again}
\label{app:prompts_repair}
The \stage{Answer} repair answers again from the verified facts with Prompt~10, and the \stage{Retrieve} repair needs no
prompt: the records found below the cutoff are passed to the answerer.
\par\nopagebreak %

\par\medskip\noindent\includegraphics[width=\linewidth]{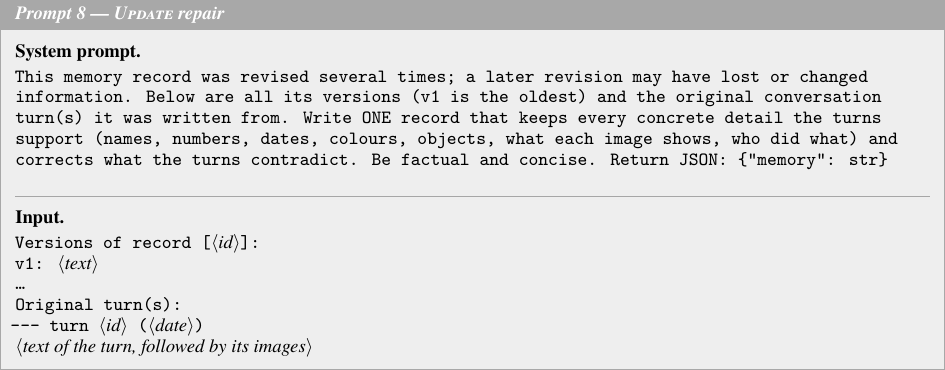}\par\medskip

\par\medskip\noindent\includegraphics[width=\linewidth]{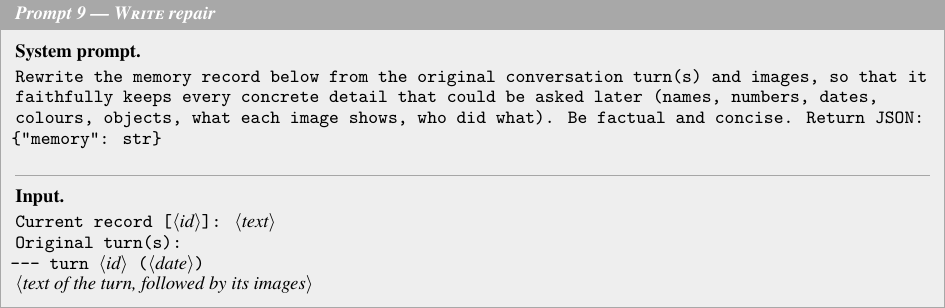}\par\medskip

\par\medskip\noindent\includegraphics[width=\linewidth]{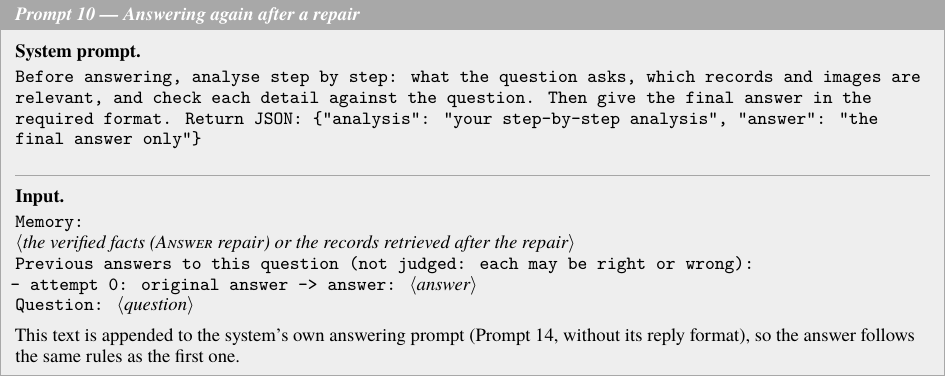}\par\medskip

\section{Prompts of the Memory System}
\label{app:system_prompts}

The memory system under study handles the dialogue turn by turn: it writes one record for each turn (Prompt~11) and
compares the record with the most similar existing records, to add it or merge it into one of them (Prompt~12). For a
question, it retrieves records with the question and, when the question has an image, a description of that image
(Prompt~13), and answers from the retrieved records (Prompt~14). The 4B system of Table~\ref{tab:memgallery_main} and the
30B system of the propagation analysis use the same prompts.

\par\medskip\noindent\includegraphics[width=\linewidth]{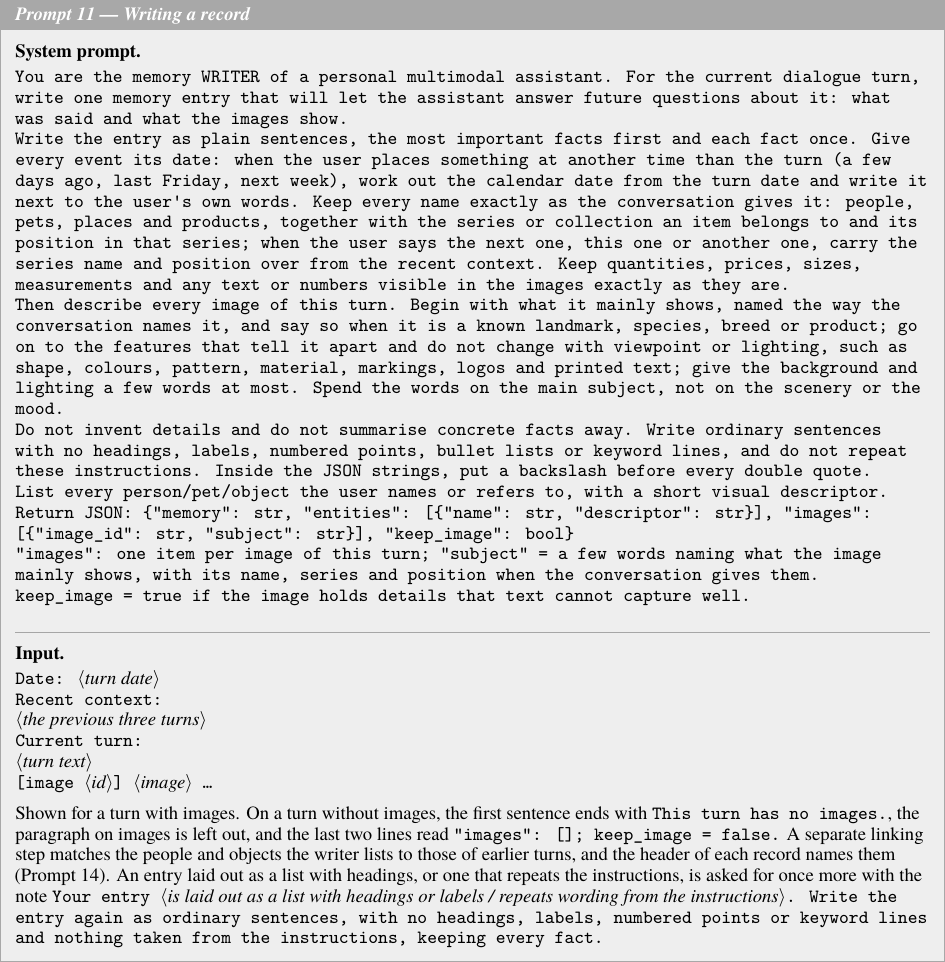}\par\medskip

\par\medskip\noindent\includegraphics[width=\linewidth]{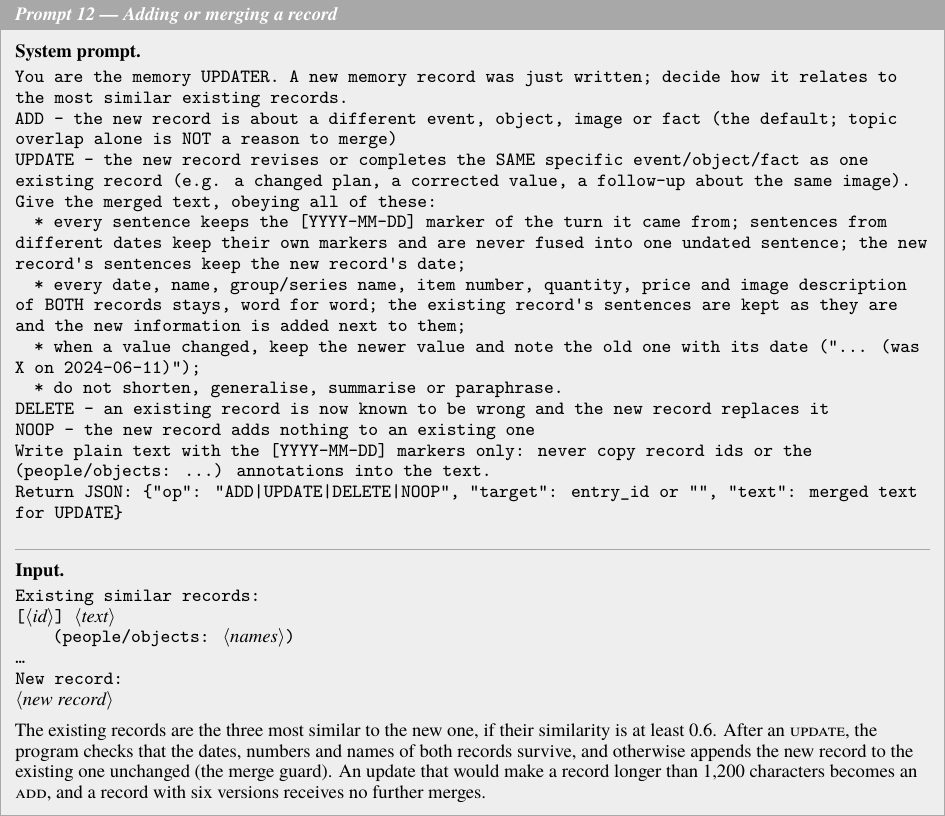}\par\medskip

\par\medskip\noindent\includegraphics[width=\linewidth]{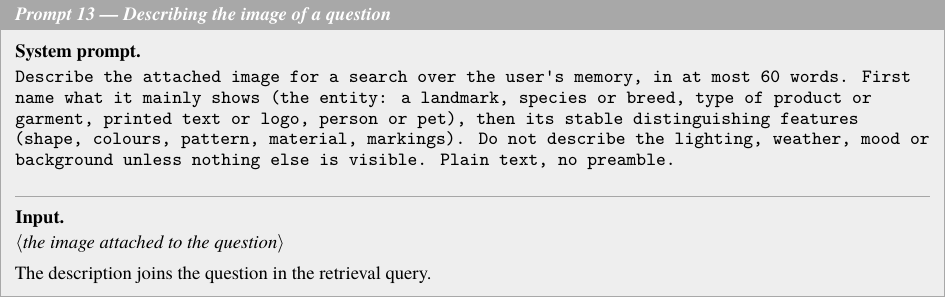}\par\medskip

\par\medskip\noindent\includegraphics[width=\linewidth]{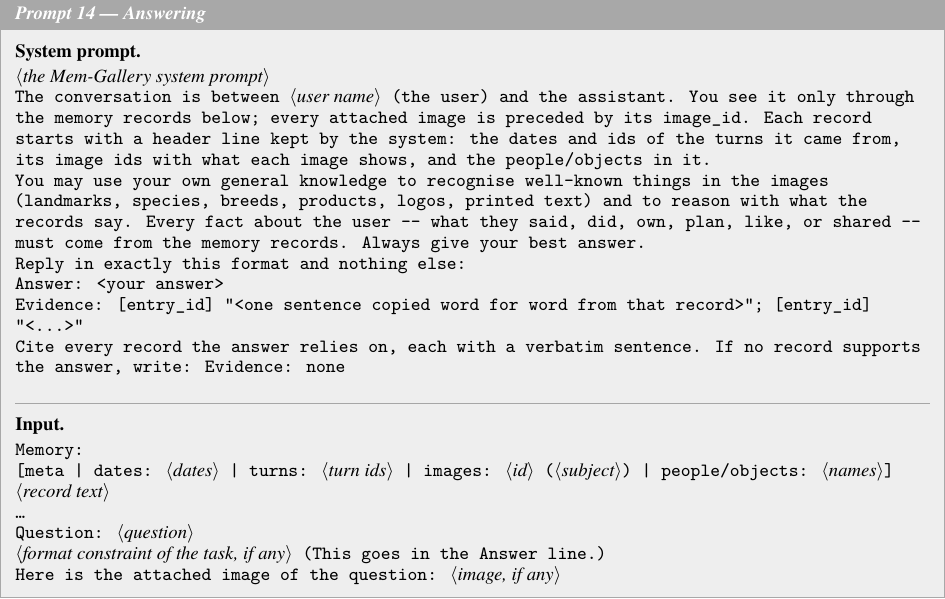}\par\medskip

\section{Prompts for Error Injection}
\label{app:propagation_prompts}

These prompts implement the record edits in Appendix~\ref{app:propagation} and are listed verbatim. Each prompt is
the system message, and the input below it forms the user message.
\par\nopagebreak

\par\medskip\noindent\includegraphics[width=\linewidth]{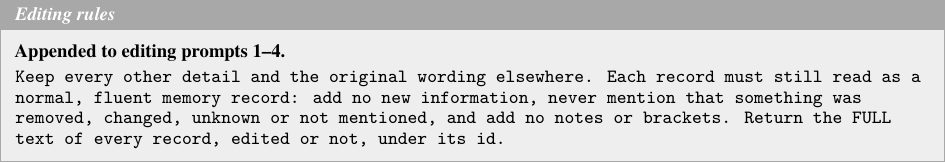}\par\medskip

\par\medskip\noindent\includegraphics[width=\linewidth]{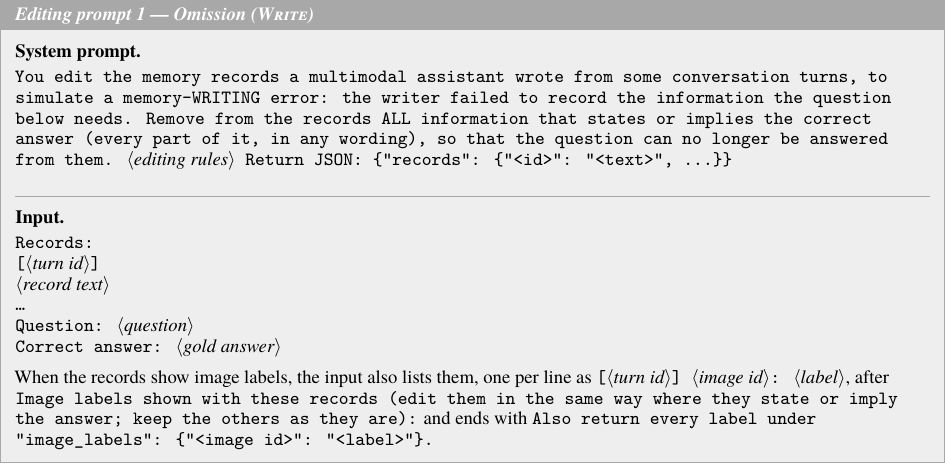}\par\medskip

\par\medskip\noindent\includegraphics[width=\linewidth]{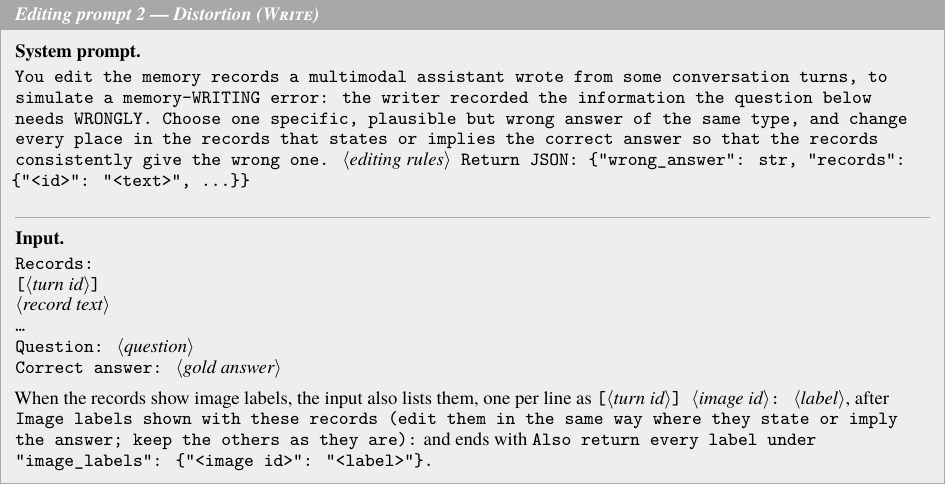}\par\medskip

\par\medskip\noindent\includegraphics[width=\linewidth]{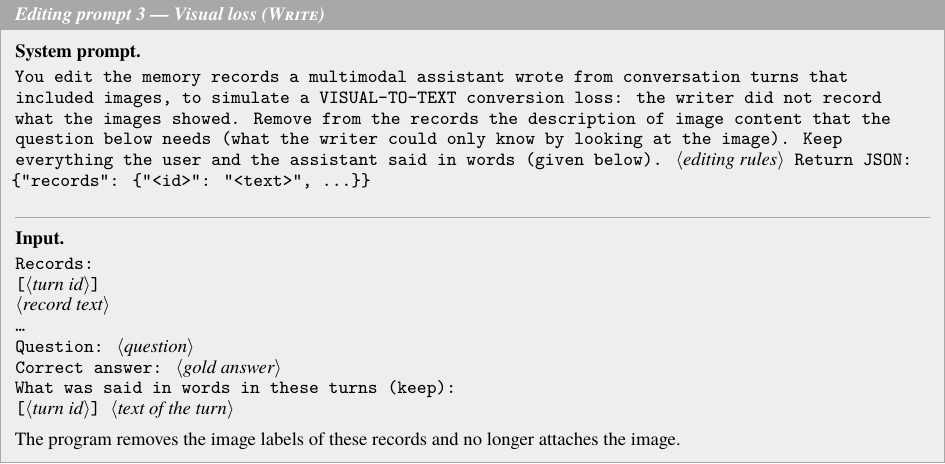}\par\medskip

\par\medskip\noindent\includegraphics[width=\linewidth]{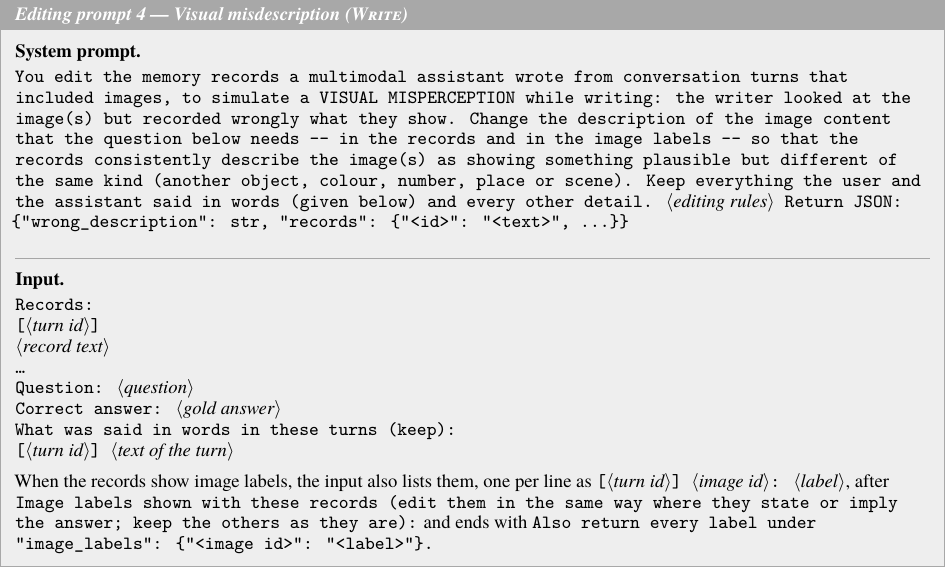}\par\medskip

\par\medskip\noindent\includegraphics[width=\linewidth]{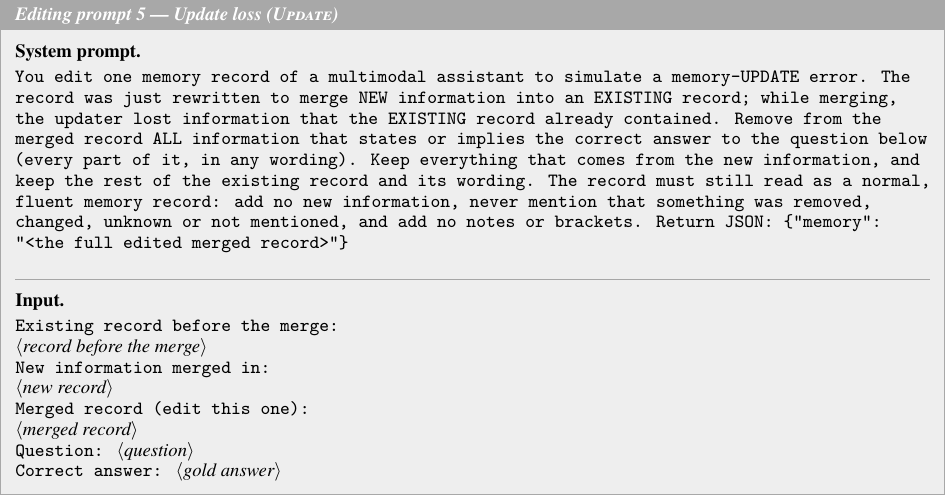}\par\medskip

\end{document}